%% file: main.tex
\documentclass[times,twocolumn,final,longtitle]{elsarticle}

\usepackage{marvosym}

\newcommand{\arxiv}{1} 

\ifdefined\arxiv
    \usepackage{medima-arxiv}
    \newcommand{\writer}[1]{~}
    \newcommand{\orcid}[1]{\href{https://orcid.org/#1}{\includegraphics[scale=0.09]{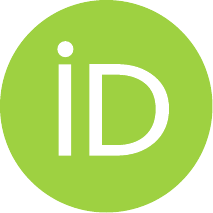}}}
    \newcommand{\mailto}[1]{\href{mailto:#1}\textsuperscript{\textsuperscript{\Letter,}} }
    \journal{}
    \date{}
    \newcommand{\red}[1]{#1}

\else
    \usepackage{medima}
    \newcommand{\writer}[1]{}
    \newcommand{\orcid}[1]{}
    \newcommand{\mailto}[1]{}
\fi

\usepackage{framed}
\usepackage{multirow}
\usepackage{multicol}
\usepackage{tabularx}
\usepackage{booktabs}
\usepackage{amssymb}
\usepackage{latexsym}
\usepackage[obeyspaces]{url}
\usepackage[dvipsnames]{xcolor}
\usepackage{rotating}
\usepackage{makecell}
\usepackage{pifont}
\usepackage{array}
\usepackage{etoolbox}

\usepackage[colorlinks=true,citecolor=blue,urlcolor=blue]{hyperref}
\AtBeginDocument{%
  \hypersetup{
    citecolor=teal,
    linkcolor=blue,   
    urlcolor=Blue}}

\newcommand{\triplet}[1]{\textlangle{\textit{#1}}\textrangle{}}

\newcommand*\OK{\ding{51}}
\newcolumntype{P}[1]{>{\centering\arraybackslash}p{#1}}
\newcommand{\teal}{\color{teal}}
\newcommand{\green}{\color{PineGreen}}

\newcommand{\team}[3]{%
    \subsubsection{{\bf #1:~}%
    {\noindent\it\teal#2}}%
    {\noindent#3}
}

\newcommand{\method}[3]{%
    \subsubsection{{\bf#1:~~}%
    {\small\teal\it#2}}%
    {\leavevmode\indent#3}
}

\preto\tabular{\setcounter{magicrownumbers}{0}}
\newcounter{magicrownumbers}

\definecolor{newcolor}{rgb}{.8,.349,.1}
\journal{Medical Image Analysis}

\begin{document}

\verso{Chinedu I. Nwoye \textit{et~al.}}

\begin{frontmatter}

\title{{\bf CholecTriplet2021: A benchmark challenge for surgical action triplet recognition}}%


\author[1]{ 
\orcid{0000-0003-4777-0857} Chinedu Innocent \snm{Nwoye} \mailto{nwoye.chinedu@gmail.com}} \corref{cor1} \cortext[cor1]{Corresponding author
}
\author[1]{Deepak \snm{Alapatt}}
\author[1]{Tong \snm{Yu}}
\author[2]{Armine \snm{Vardazaryan}}

\author[3]{Fangfang \snm{Xia}}
\author[3]{Zixuan \snm{Zhao}}

\author[4]{Tong \snm{Xia}}
\author[4]{Fucang \snm{Jia}}

\author[6]{\orcid{0000-0001-5990-4061} Yuxuan \snm{Yang}}
\author[6]{\orcid{0000-0001-6503-981X} Hao \snm{Wang} }

\author[7]{Derong \snm{Yu}}
\author[7]{\orcid{ 0000-0003-4173-0379} Guoyan \snm{Zheng} }
\author[8]{Xiaotian \snm{Duan}}
\author[8]{Neil \snm{Getty}}
\author[9]{Ricardo \snm{Sanchez-Matilla}}
\author[9]{Maria \snm{Robu}}
\author[10]{Li \snm{Zhang}}
\author[10]{Huabin \snm{Chen}}
\author[11]{Jiacheng \snm{Wang}}
\author[11]{Liansheng \snm{Wang}}
\author[12]{Bokai \snm{Zhang}}
\author[13]{\orcid{0000-0003-2570-1834} Beerend \snm{Gerats} }

\author[14]{Sista \snm{Raviteja}}
\author[14]{\orcid{0000-0003-3402-3852} Rachana \snm{Sathish}}
\author[7]{\orcid{0000-0002-6843-3668} Rong \snm{Tao} }
\author[16]{Satoshi \snm{Kondo}}
\author[17]{Winnie \snm{Pang}}
\author[23]{Hongliang \snm{Ren}}
\author[13]{\orcid{0000-0001-6899-1048} Julian Ronald \snm{Abbing} }
\author[12]{Mohammad Hasan \snm{Sarhan}}
\author[18]{Sebastian \snm{Bodenstedt}}
\author[18]{Nithya \snm{Bhasker}}
\author[19,20,21]{\orcid{0000-0003-1616-2234} Bruno \snm{Oliveira}}
\author[19,20,21]{\orcid{0000-0001-8407-3255} Helena R. \snm{Torres}}
\author[6]{Li \snm{Ling}}
\author[222]{Finn \snm{Gaida}}
\author[222]{Tobias \snm{Czempiel}}


\author[19]{\orcid{0000-0002-4196-5357} João L. \snm{Vilaça}} 
\author[19]{\orcid{0000-0002-1995-7879} Pedro \snm{Morais}}
\author[21]{\orcid{0000-0001-6703-3278} Jaime \snm{Fonseca}}
\author[13]{Ruby Mae \snm{Egging}}
\author[13]{Inge Nicole \snm{Wijma}}
\author[10]{Chen \snm{Qian}}
\author[10]{Guibin \snm{Bian}}
\author[10]{Zhen \snm{Li}}
\author[14]{Velmurugan \snm{Balasubramanian}}
\author[14]{\orcid{0000-0001-9046-149X} Debdoot \snm{Sheet}}
\author[9]{Imanol \snm{Luengo}}

\author[6]{\orcid{0000-0002-6193-7054} Yuanbo \snm{Zhu} }
\author[6]{\orcid{0000-0002-8384-1950} Shuai \snm{Ding}}

\author[12]{Jakob-Anton \snm{Aschenbrenner}}
\author[13]{Nicolas Elini \snm{van der Kar}}
\author[17]{Mengya \snm{Xu}}
\author[17]{Mobarakol \snm{Islam}}
\author[17]{Lalithkumar \snm{Seenivasan}}
\author[18]{Alexander \snm{Jenke}}
\author[9,24]{Danail \snm{Stoyanov}}

\author[2,22]{Didier \snm{Mutter}}
\author[1,25]{\orcid{0000-0001-7288-3023} Pietro \snm{Mascagni}}
\author[1,2,22]{\orcid{0000-0001-7451-7545} Barbara \snm{Seeliger}}
\author[2,22]{Cristians \snm{Gonzalez}}

\author[1,2]{\orcid{0000-0002-5010-4137} Nicolas \snm{Padoy}}

\address[1]{ICube, University of Strasbourg, CNRS, France}
\address[2]{IHU Strasbourg, France}
\address[3]{Department of Computer Science, University of Chicago, US}
\address[4]{Lab for Medical Imaging and Digital Surgery, Shenzhen Institute of Advanced Technology, Chinese Academy of Sciences, China}
\address[6]{School of Management, Hefei University of Technology, Hefei, China}
\address[7]{Institue of Medical Robotics, Shanghai Jiao Tong University, Shanghai, China}
\address[8]{Argonne National Laboratory, 9700 S Cass Ave, Lemont, IL 60439, US} 
\address[9]{Digital Surgery, a Medtronic Company, London, UK}
\address[10]{Institute of Automation, Chinese Academy of Sciences, China}
\address[11]{Department of Computer Science at School of Informatics, Xiamen University, Xiamen, China}
\address[12]{Johnson \& Johnson}
\address[13]{Meander Medical Centre, The Netherlands}
\address[14]{Indian Institute of Technology Kharagpur, India}
\address[16]{Muroran Institute of Technology, Japan}
\address[17]{Department of Biomedical Engineering, National University of Singapore, Singapore}
\address[18]{Department for Translational Surgical Oncology, National Center for Tumor Diseases Partner Site Dresden, Germany}
\address[19]{2Ai School of Technology, IPCA, Barcelos, Portugal.}
\address[20]{Life and Health Science Research Institute (ICVS), School of Medicine, University of Minho, Braga, Portugal. }
\address[21]{Algoritimi Center, School of Engineering, University of Minho, Guimeraes, Portugal.}
\address[222]{Technical University of Munich, Germany}
\address[23]{Department of Electronic Engineering, The Chinese University of Hong Kong, Hong Kong}
\address[24]{Wellcome/EPSRC Center for Interventional and Surgical Science, University College London, UK}
\address[25]{Fondazione Policlinico Universitario Agostino Gemelli IRCCS, Rome, Italy}
\address[22]{University Hospital of Strasbourg, France}

\received{: }
\finalform{:}
\accepted{:}
\availableonline{:}
\communicated{:}

\input{main/00-abstract}

\end{frontmatter}


\input{main/01-introduction}
\input{main/02-literature}
\input{main/04-challenge}

\input{main/05-methods}

\input{main/06-evaluation}

\input{main/07-results}
\input{main/08-conclusion}

\input{main/09-acknowledgement}

\bibliographystyle{model2-names.bst}\biboptions{authoryear}
\bibliography{main}

%

\ifdefined\arxiv
    \input{credit}
\fi
\end{document}

%% file: main/00-abstract.tex
\begin{abstract}
Context-aware decision support in the operating room can foster surgical safety and efficiency by leveraging real-time feedback from surgical workflow analysis. Most existing works recognize surgical activities at a coarse-grained level, such as phases, steps or events, leaving out fine-grained interaction details about the surgical activity; yet those are needed for more helpful AI assistance in the operating room.
Recognizing surgical actions as triplets of \triplet{instrument, verb, target} combination delivers comprehensive details about the activities taking place in surgical videos.
This paper presents {\it CholecTriplet2021}: an endoscopic vision challenge organized at MICCAI 2021 for the recognition of surgical action triplets in laparoscopic videos. 
The challenge granted private access to the large-scale {\it CholecT50} dataset, which is annotated with action triplet information.
In this paper, we present the challenge setup and assessment of the state-of-the-art deep learning methods proposed by the participants during the challenge.
A total of 4 baseline methods from the challenge organizers and 19 new deep learning algorithms by competing teams are presented to recognize surgical action triplets directly from surgical videos, achieving mean average precision (mAP) ranging from 4.2\% to 38.1\%.
This study also analyzes the significance of the results obtained by the presented approaches, performs a thorough methodological comparison between them, in-depth result analysis, and proposes a novel ensemble method for enhanced recognition.
Our analysis shows that surgical workflow analysis is not yet solved, and also highlights interesting directions for future research on fine-grained surgical activity recognition which is of utmost importance for the development of AI in surgery.
\end{abstract}

\begin{keyword}
\vspace{-0.33in}
\KWD Surgical activity recognition\sep tool-tissue interaction\sep CholecT50\sep laparoscopic video \sep action recognition.
\end{keyword}

%% file: main/01-introduction.tex
\section{Introduction}
\begin{figure}[ht]
    \includegraphics[width=1.0\linewidth]{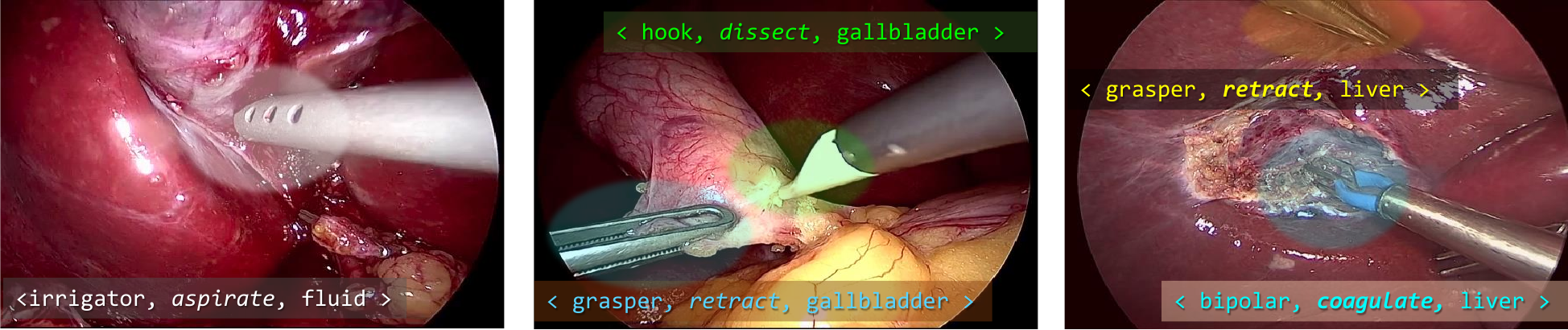} 
    \caption{A cross-section of CholecT50 dataset for the CholecTriplet2021 challenge. Represented surgical action triplets are illustrated for different time-points during a laparoscopic cholecystectomy.}
    \label{fig:data}
\end{figure}

Activity recognition is the basis for the development of many potential applications in health, surveillance, manufacturing, sports, etc \citep{avci2010activity}.
In surgery, it can be leveraged to provide intra-operative context-awareness and decision support \citep{maier2017surgical}.
This could augment surgeons' capabilities, fostering safety and efficiency in the operating room (OR) \citep{vercauteren2019cai4cai}.

Despite the vast literature in medical computer vision, a majority of the works tackle activity recognition at a very coarse-grained level such as phase \citep{twinanda2016endonet}, which does not provide an accurate picture of activities taking place. As an example, the {\it calot triangle dissection} phase in cholecystectomy contains a multitude of finer actions that would be relevant to recognize.
The finer division of activity recognition such as step \citep{lecuyer2020assisted}, gesture \citep{dipietro2019segmenting} or action \citep{wagner2021comparative} recognition leaves out details about the anatomy and thus, does not provide comprehensive details to describe the tool-tissue interaction.
The detection of the used surgical instruments and their target anatomy as well as their fine-grained interaction details is necessary for the development of artificial intelligence (AI) that is safe for the patient \citep{nwoye2021deep}.
This opens a vista of opportunities to develop methods that recognize the elements involved in tool-tissue interaction while accounting for the relationship and multiple instances.

In general computer vision, human activity is modeled as triplets \triplet{human, verb, object} providing full-scale and expressive details on Human-object interaction (HOI) \citep{chao2015hico}. Translating this formalism to surgical vision, however, was not achieved until very recently, when \cite{nwoye2020recognition} presented the video recognition of surgical activities as triplets of the used instruments, actions performed, and the underlying target anatomy.

We present a critical study on surgical action triplet recognition in the form of a challenge termed {\bf CholecTriplet2021}\footnote{\url{https://cholectriplet2021.grand-challenge.org/}}, to pave the way for research targeting fine-grained and detailed recognition of surgical activities from videos.
This international challenge was organized as part of the Endoscopic Vision (EndoVis) Grand Challenge \citep{speidel2021endoscopic} and hosted at MICCAI 2021.
The challenge presented promising technologies for the detailed understanding of tool-tissue interactions in minimally invasive surgery.
A total of 19 teams participated, the highest record since the inception of the EndoVis Grand Challenge series.

The challenge provided a platform for the scientific community to perform comparative benchmarking and validation of endoscopic vision algorithms in a promising direction in surgical activity recognition.
We provided private access to a high-quality large-scale surgical action triplet dataset, {\it CholecT50} \citep{nwoye2021rendezvous}, for both method development and validation.
In addition to these contributions brought by the event itself, the challenge report presented here offers contributions of its own. After individual descriptions for each approach, we highlight several trends in an in-depth methodological comparison, providing a comprehensive overview of possible strategies for tackling the surgical action triplet problem. To further facilitate future research efforts, we also compile the implementation details of all featured submissions. We then set a new upper baseline (+4.3\% AP to challenge methods) for the surgical action triplet recognition problem, by proposing a simple but effective algorithm ensembling models featured in the challenge. Finally, results are analyzed in depth: our quantitative analysis considers multiple metrics to cover all aspects of the triplet recognition problem. A rich selection of qualitative results is presented as well, to better understand the behavior of all the methods.

The paper is organized as follows: we position our work with respect to the related literature in the next section. Afterward, we present the challenge overview and setup including the used dataset and a summary of participation. This is followed by the methods presented at the challenge including the evaluation protocols, results, and extensive analysis of their performance. We conclude by highlighting the benefits of this study, insights, and prospects.

%% file: main/02-literature.tex
\section{Related work}
The {CholecTriplet2021} challenge relates to several research topics, for which we present the relevant literature in the following paragraphs.

\subsection{Activity recognition}
One of the key concerns of computer vision is the recognition of human activities; a task for which a wide variety of approaches has been proposed, both in medical and non-medical domains. In those approaches, the definition of this task can be more or less granular, which is a key factor in determining its difficulty as well as its utility. Early activity recognition work in general computer vision involved broad, coarse-grained classification tasks: the 2012 version of the PASCAL VOC \citep{everingham2015thepascal} challenge proposed an action classification task on static images with 10 classes. HMDB-51 \citep{kuehne2011hmdb} and UCF-101 \citep{soomro2012ucf101}  are collections of realistic action video clips extracted from YouTube. As datasets expanded, the number of classes increased; classes became more diverse but also finer-grained. For example, Kinetics \citep{jo2017quo} features three classes related to cycling (\textit{"riding a bike", "riding a mountain bike", "falling off a bike"}) while HMDB-51 only contains one \citep{kuehne2011hmdb}. Generally speaking, the advantage of having more granular classes is that they enable more detailed and informative descriptions.

In surgical computer vision, fine-grained activity descriptions are particularly valuable: as key components in concepts for context-aware surgical systems \citep{maier2017surgical,maier2017surgical2}, activity recognition algorithms tie into clinical outcomes. However, descriptions achieved by surgical vision algorithms have historically been coarse. The task of surgical phase recognition \citep{twinanda2016endonet,dergachyova2016automatic,funke2018temporal,yu2019learning,zisimopoulos2018deepPhase,hajj2018monitoring,czempiel2020tecno}, one of the main research topics in this area, breaks down an entire surgery into a small number of chunks, each with a broadly defined role within the surgical procedure. Since phases can last several minutes, they may contain many individual actions; this lack of detail is ultimately what limits the utility of the phase information. For this reason, other studies addressed more granular classes: subdivisions of phases into steps can be found in \cite{ramesh2021multi,lecuyer2020assisted} and recognition of action verbs has are explored in \cite{rupprecht2016sensor,khatibi2020proposing}. In the previous edition of EndoVis, an action recognition task was featured \citep{wagner2021comparative}, modeling tool-tissue interaction to a certain degree: each class was defined by one verb.

Overall, a shift towards finer-grained activity recognition has become a major focus in recent research on activity recognition, especially for surgery.

\subsection{Action triplet recognition}
The finest level of granularity in visual activity understanding is currently achieved by decomposing actions into their individual components: who performs the action, what the action is, and what the action is performed on. This decomposition into action triplets of \textit{(subject, verb, object)} is central in HOI \citep{chao2015hico} studies. \cite{mallya2016learning} used CNN features extracted from humans and objects detected in frames. \cite{yuwei2018learning} proposed a multi-stream architecture to model spatial relationships. \cite{gkioxari2018detecting}'s method was built around a FasterRCNN object detector to locate humans. \cite{qi2018learning} introduced the Graph Parsing Neural Network (GPNN), representing human-object interactions with the adjacency matrix and node labels of a graph.

In surgical computer vision, action triplets \citep{katic2014knowledge} in the form of \triplet{surgical tool/instrument, action verb, target anatomy} are used to describe tool-tissue interactions (TTI) \citep{nwoye2021deep}. Early works used them as auxiliary annotations to improve surgical phase recognition \citep{katic2014knowledge,katic2015lapontospm}. The first method to actually perform action triplet recognition from videos was introduced by \cite{nwoye2020recognition} as the Tripnet, featuring verb and target detection using instrument class activation guidance, as well as triplet association using a 3D interaction space. A new model \citep{nwoye2021rendezvous} with more advanced modeling of interactions between triplet components using an attention mechanism was later developed. \cite{xu2021learning} proposed a cross-domain method for automatic surgical captions that capture semantic relationships, similarly to action triplets.
\begin{figure*}[ht]
\centering
    \includegraphics[width=.75\linewidth]{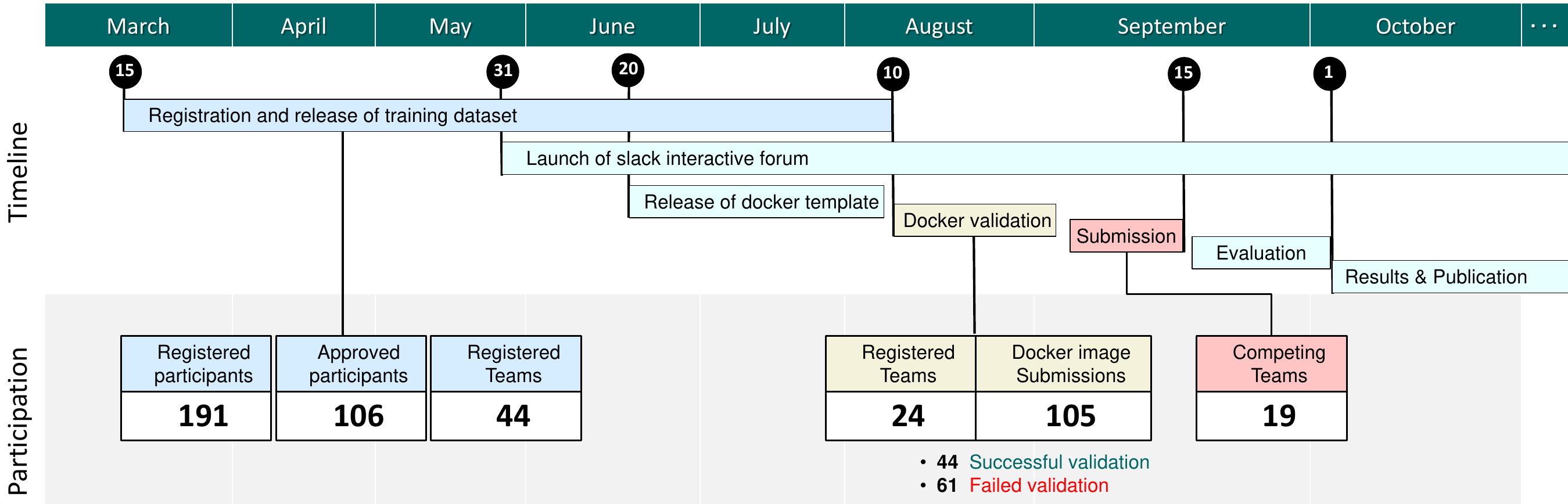} 
    \caption{CholecTriplet2021 challenge timeline and activities.}
    \label{fig:timeline}
\end{figure*}

\subsection{Action triplet datasets}
The study of action triplet recognition requires datasets annotated with triplet components. In general computer vision, an early example is the HICO dataset \citep{yuwei2015hico}. CAD-120 \citep{koppula2013learning} is annotated with object affordances. V-COCO \citep{sadhu2021visual}, as an extension of the widely used MS-COCO \citep{tsung2014microsoft}, added visual semantic role labels; the provided bounding box annotations also enabled HOI spatial detection. HICO later received an update in the form of HICO-DET \citep{yuwei2018learning}, similarly incorporating bounding boxes. HCVRD \citep{zhuang2018hcvrd} was proposed as a benchmark for detecting human-centered visual relationships, which includes action verbs among other types. Ambiguous-HOI \citep{yonglu2020detailed} collects difficult examples from several of the datasets previously mentioned, in order to form a challenging HOI benchmark.

In the surgical domain, datasets offering action triplet annotations describing tool-tissue interaction are much more recent. In a challenge organized by \cite{wagner2021comparative}, a cholecystectomy video dataset with annotations for four surgical action verbs was provided. The SARAS-ESAD dataset \citep{vivek2021thesaras} featured more refined classes with actions described by both the verb and the anatomy; bounding boxes for 21 action classes were provided as well. The first dataset with full annotations for each triplet component was introduced in \cite{nwoye2020recognition} as CholecT40. The expanded version renamed CholecT50 \citep{nwoye2021rendezvous} is employed in this challenge. Another dataset used in \cite{xu2021learning} incorporated surgical captions; despite not explicitly following the action triplet formalism, the level of detail offered is similar.

\subsection{Endoscopic vision challenges}
As a relatively new problem, surgical action triplet recognition has only received little attention despite its potential. This is a major motivating factor for \textit{CholecTriplet2021}: over the past few years, challenges have played an important role in the surgical computer vision community due to the exposure they can bring to interesting research topics, as well as their ability to introduce and compare a wide variety of original methods. For instance, the M2CAI 2016 challenge \citep{stauder2016thetum} featured two tasks - surgical phase recognition and tool presence detection for cholecystectomy. The first edition of the CATARACTS challenge \citep{hajj2019cataracts}, involved tool recognition on cataract surgery videos; later editions became part of the EndoVis (Endoscopic Vision) challenge series. Each iteration of EndoVis featured multiple sub-challenges, some of which focused on surgical activity understanding: segmentation and tracking of tools for colorectal surgery videos (2015), phase recognition for colorectal surgery using video and sensor data (2017), and finally, phase, tool and action recognition for cholecystectomy videos \citep{wagner2021comparative}.

Outside of EndoVis subchallenges, the SARAS-ESAD \citep{vivek2021thesaras} organized within the MIDL 2020 challenge presents a benchmark for surgical action detection using a large-scale video dataset (ESAD) offering another level of granularity (\triplet{verb, target} as a single label) as well as bounding boxes for action localization.

%% file: main/04-challenge.tex
\section{CholecTriplet challenge}

The goal of this challenge is to assess AI solutions for fine-grained surgical activity recognition. This recognition is modeled as a triplet \citep{nwoye2020recognition}, with the following notation: \triplet{instrument, verb, target}

\subsection{Task}
The task is to develop a machine learning method to recognize these triplets directly from unseen surgical videos.
In modeling surgical action triplet, a prevailing problem to tackle is the simultaneous detection of the correct instruments, verbs, and targets in every image frame and resolving their association. This is challenging since the involvement of a component in a triplet can be visually subtle.
Hence, the challenge also assesses the detection of these individual components for a more insightful analysis of a model's understanding of the triplet's composition.

\subsection{Dataset }
The dataset used for the challenge is \textit{CholecT50} \citep{nwoye2021rendezvous}, which consists of 50 video recordings of laparoscopic cholecystectomy that have been annotated with the binary presence of action triplets. At 1 frame per second (fps), the dataset reaches a total of 100.9K frames and 161K triplet instance labels.
{\red
CholecT50 consists of 100 action triplet classes composed from 6 instruments (\textit{grasper, bipolar, hook, scissors, clipper, irrigator}), 10 verbs (\textit{grasp, retract, dissect, coagulate, clip, cut, aspirate, irrigate, pack, null}), and 15 targets (\textit{gallbladder, cystic-duct, cystic-artery, blood-vessel, fluid, abdominal-wall or cavity, liver, omentum, peritoneum, gut, specimen-bag, null}). 
}
The triplet labels we provided in form of a single binary presence for each triplet class \triplet{instrument, verb, target} taken as a whole. Binary labels for each component of the triplet were also provided.

On the data split, 45 out of 50 videos in the dataset were released to the participants for training and validation; those videos were part of the publicly released \textit{Cholec80} dataset \cite{twinanda2016endonet}.
The remaining 5 videos, which were not public, were withheld from the participants as the testing set.

\begin{table*}[htp]
    \centering
    \caption{Demography of final participating teams.$^\dag$}
    \label{tab:demography}
    \resizebox{0.8\textwidth}{!}{%
    \begin{tabular}{c|c|c|c|c|c|c|c|c|c}
        \hline
        \multicolumn{4}{c|}{Asia} & \multicolumn{5}{c|}{Europe} & \multicolumn{1}{c}{North America}\\\hline
        China & Singapore & Japan & India & Germany & UK & Netherlands & Portugal & France & USA \\\hline
8 & 2 & 1 & 1 & 3 & 2 & 1 & 1 & 1 & 3 \\ \hline
        \multicolumn{9}{l}{$\dag$ \scriptsize{Some teams have affiliations in multiple countries}.}\\
    \end{tabular}
    }
\end{table*}

\subsection{Challenge design }

The challenge was conducted as part of EndoVis grand-challenge \citep{speidel2021endoscopic} in MICCAI 2021. The challenge proposal went through two rounds of open review between November 2020 and February 2021.
By early March 2021, a call for participation was circulated and the challenge officially started on March 15, 2021, with a public release of the training data.
Participation took place by web registration and signing a non-disclosure contract on the use of the challenge dataset.

By the end of May, a Slack\footnote{\url{https://www.slack.com}} channel was created for the registered participants and their teams.
A blog was provided with snippets of code and instructions for getting started with the challenge. Reference to the baseline methods and code for some specialized functions were also provided. 
A Docker\footnote{\url{https://hub.docker.com}} submission template, development guide, and evaluation metrics were provided to the participants during their method development phase.
Participating teams were allowed to develop novel methods, fine-tune a state-of-the-art method, or improve on existing solutions.
Entrants were allowed to pre-train their model on any third-party publicly available dataset.
Developed methods were intended to take sequential image frames from videos, process, and return a vector probability for the 100 triplet classes for each image.
Due to the possibility of multiple instances of the triplets, both the triplets and their three components were cast as a multi-label classification problem.
The output probability vectors for the three components of the triplets were derived using a filtering algorithm proposed in \cite{nwoye2021rendezvous}.

By the terminal point of the development phase, a validation process was initiated for users to ascertain that their Docker container had been built with the correct input/output format and was able to run without run-time errors on a set of randomly selected images from the training set. Teams were allowed to validate multiple times until an error-free Docker was obtained, but within a time-bound of three weeks which lasted till Sept 5, 2021.
The final submission was performed once using only a validated Docker image container. These submissions were run and evaluated on the hidden test set by the challenge organizers with the outcome presented at the MICCAI 2021 satellite event.
The challenge timeline is presented in Fig. \ref{fig:timeline}.

We also provide a one-month post-challenge window (November 15 - December 15, 2021) during which all teams could re-evaluate their updated model if their prior submissions had inaccuracies.

\subsection{Method submission}
All Docker submissions were collected as saved Docker images uploaded to Dropbox\footnote{\url{https://www.dropbox.com}}. Only Dockers strictly adhering to a predefined Docker format and input/output requirements were considered as valid submissions. In practice, this involved passing several automatic checks including ensuring compatibility with a defined directory structure and naming format. In terms of output format, automatic checks were conducted to ensure that an output probability value between 0 and 1 was written for each input frame and consequently each available video. Additionally, each submitted method was required to pass a causality check on a predefined, hidden subset of frames to ensure that future frames were not utilized. During the validation phase, participants were expected to make corrections to the submissions based on automatically generated feedback that was shared by email to ensure that only validated Docker images were submitted for final evaluation.

\subsection{Awards }
{The winning submission earned an NVIDIA GPU. Monetary prizes were also awarded to the top 3 competing teams.}

\subsection{Participation statistics}
As presented in Fig. \ref{fig:timeline}, an initial 44 teams registered by the end of the enrollment window, accounting for 106 approved individual participants out of 191 recorded registrations. During the Docker validation phase, 24 teams submitted a total of 105 Docker containers which were evaluated with a 41.9\% success rate. On the deadline of September 15, 2021, a total of 19 teams' submissions were received for final evaluation. A baseline submission from the organizers brings this to a total of 20 teams. {\red The organizer's submission consists of four different baseline methods.} The participating teams were drawn from 10 countries across 3 continents. The demography of the final participating teams is presented in Table \ref{tab:demography}.

\begin{table*}[t]
    \centering
    \caption{~A complete cross-view of the MICCAI EndoVis CholecTriplet 2021 participating teams including their affiliations and presented methodologies.}
    \label{tab:teams}
    \setlength{\tabcolsep}{5pt}
    \resizebox{\textwidth}{!}{%
    \begin{tabular}{@{}rlll@{}}
        \toprule
        & {\bf Team } & {\bf Affiliation(s)} & {\bf Method} \\ \toprule
        1 & 2Ai & Applied Artificial Intelligence Laboratory, Portugal & \makecell[tl]{Surgical video analysis using an ensemble of multi-task recurrent\\ convolutional networks (versions 1 \& 2)} \\ \midrule
        2 & ANL-Triplet & Argonne National Laboratory, USA & \makecell[tl]{ANL-Triplet: Exploiting temporal information for triplet recognition} \\\midrule
        3 & Band of Broeders &  \makecell[tl]{Meander Medical Centre, The Netherlands \\ Johnson \& Johnson, USA} & YoloV5 for surgical action triplet detection \\\midrule
        4 & CAMMA & University of Strasbourg, France (Organizers) & MTL Baseline, Tripnet, Attention Tripnet, and Rendezvous \\\midrule
        5 & CAMP & Technical University of Munich, Germany & \makecell[tl]{EndoVisNet: Phase-guided temporal endoscopic action triplet\\classification}\\\midrule
        6 & Casia Robotics & Institute of Automation, Chinese Academy of Sciences, China & \makecell[tl]{Triplet translation embedding network with coordinate attention}\\\midrule
        7 & Ceaiik & Indian Institute of Technology Kharagpur, India & \makecell[tl]{Spatio-temporal learning of action triplets in surgical videos}\\\midrule
        8 & CITI SJTU & Shanghai Jiao Tong University, China & \makecell[tl]{Action triplet recognition via convolutional LSTMs and multi-task\\learning}\\\midrule
        9 & Digital Surgery & \makecell[tl]{Digital Surgery, a Medtronic Company, London, UK\\
        Wellcome/EPSRC Center for Interventional and Surgical \\Science, University College London, UK} & \makecell[tl]{TE-TAR: Temporal ensemble triplet action recognition} \\\midrule
        10 & Lsgroup & Xiamen University, China & \makecell[tl]{Feature fusion and weak locational information calculating in triplet\\ classification multi-task} \\\midrule
        11 & HFUT-MedIA & Hefei University of Technology, China & \makecell[tl]{COEMNet: Correlation embedded multi-task network} \\\midrule
        12 & HFUT-NUS & \makecell[tl]{National University of Singapore, Singapore\\ Heifei University of Technology, China} & Multi-task learning based surgical interaction triplet recognition\\\midrule
        13 & J\&M & \makecell[tl]{Johnson \& Johnson, US \\ Meander Medical Centre, The Netherlands} & Surgical action triplet recognition with Efficient-MSTCN\\\midrule
        14 & Med Recognizer & Chinese University of Hong Kong, China & \makecell[tl]{Surgical action triplet recognition via temporal memory relation\\gradient network} \\\midrule
        15 & MMLAB & National University of Singapore, Singapore & Temporal triplet net for triplet presence detection in surgical videos \\\midrule
        16 & NCT-TSO & \makecell[tl]{National Center for Tumor Diseases\\Partner Site Dresden, Germany} & Multi-task learning framework for action triplet recognition\\\midrule
        17 & SIAT CAMI & \makecell[tl]{Shenzhen Institute of Advanced Technology,\\ Chinese Academy of Sciences, China} & \makecell[tl]{Multi-task mutual channel recurrent net for fine-grained surgical\\ triplet recognition} \\\midrule
        18 & SJTU-IMR & Shanghai Jiao Tong University, China & \makecell[tl]{Tracking surgical actions with transformers and action label-guided\\ fine-grained information aggregation} \\\midrule
        19 & SK & Muroran Institute of Technology, Japan  & \makecell[tl]{Action triplet recognition with weakly-supervised attention of \\surgical instruments} \\\midrule
        20 & Trequartista & University of Chicago, USA & \makecell[tl]{Phase-aware multitasking action recognition model with adjustment\\for low-data triplet classes} \\ \bottomrule
    \end{tabular}
    }
\end{table*}

%% file: main/05-methods.tex
\section{Methods}

\subsection{Baseline methods}
Four baseline methods are provided by the challenge organizers ({\bf Team CAMMA}): (1) MTL baseline \citep{nwoye2020recognition}, (2) Tripnet \citep{nwoye2020recognition}, (3) Attention Tripnet \citep{nwoye2021rendezvous}, and (4) Rendezvous \citep{nwoye2021rendezvous}. We present a brief overview of the baseline models below:

\subsubsection{MTL baseline}
The Multi-Task Learning baseline \citep{nwoye2020recognition} is built around a ResNet-18 backbone, which serves as the common visual feature extractor for three separate branches. Each branch is a 3-layer (2 convolutional, 1 fully connected) neural network responsible for recognizing one of the triplet components. The instrument branch differs from the other two with the addition of Global Max Pooling (GMP). A final fully-connected layer is used to combine the three components' prediction into a final triplet prediction.

\subsubsection{Tripnet}
\begin{figure}[!htpb]
\centering
    \includegraphics[width=1.0\columnwidth]{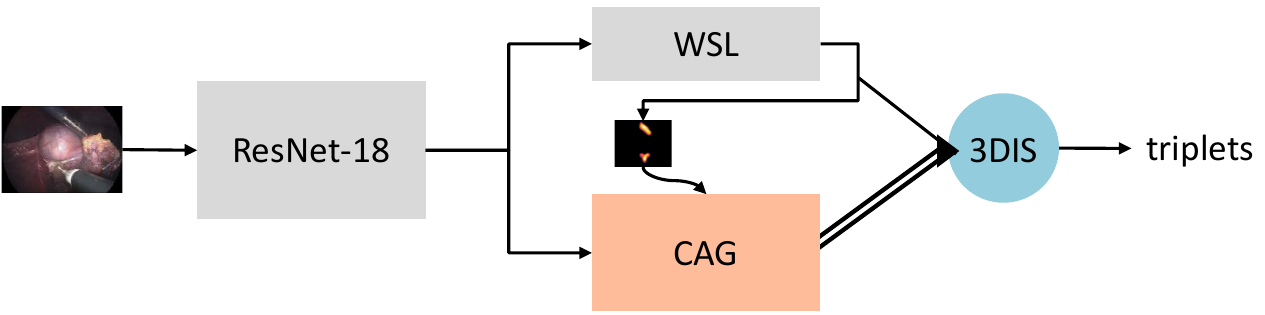} 
    \caption{Overview of Tripnet: a baseline method for action triplet recognition built on ResNet-18 backbone, weakly supervised localization (WSL) of surgical instruments, class activation guide (CAG) for verb-target recognition, and 3D interaction space (3DIS) for triplet association.}
    \label{fig:tripnet}
\end{figure}

The Tripnet \citep{nwoye2020recognition} also relies on multi-task learning from a ResNet-18 backbone but provides a stronger baseline thanks to its two characteristics. The first one is the Class Activation Guide (CAG): here the instrument branch, called the Weakly Supervised Localization (WSL) module, generates per-instrument Class Activation Maps, which are forwarded to a subnetwork in charge of verb and target detections as shown in Fig. \ref{fig:tripnet}. These maps are concatenated to the verb and target features. This additional information on instrument positions, in the form of concatenated instrument activation maps, provides clues for better detection of verbs and targets, since those are located at the tooltips.

The second innovation of the Tripnet is the 3D Interaction Space (3DIS), which associates the triplet components. Log probabilities for each component are projected into a 3D grid of dimensions $n_{I} \times n_{V} \times n_{T}$ (number of instruments, verbs, and targets, respectively) by a trainable function. Each point in the 3D grid represents the probability of the triplet, as estimated by the 3DIS function.

\subsubsection{Attention Tripnet}
The Attention Tripnet \citep{nwoye2021rendezvous} enhances the previous Tripnet, by replacing the CAG with an attention model, the Class Activation Guided Attention Mechanism (CAGAM). While the CAG only proceeds by feature concatenation, the CAGAM computes additive enhancements for verb and target class maps based on the instrument maps. Those enhancements are obtained by two different attention mechanisms: a channel (instrument type) attention for verbs and a position attention for targets. Using attention in this manner explicitly steers the verb and target detections towards the correct locations, much more strongly than the concatenation in the Tripnet's CAG.

\subsubsection{Rendezvous}
Rendezvous \citep{nwoye2021rendezvous}, or simply RDV, is the strongest baseline model. In contrast to the Attention Tripnet, triplet association is performed using a Transformer-like model built around a Multi-Head of Mixed Attention (MHMA), instead of the 3DIS. The MHMA operates on a set of four features: \textbf{$H_{IVT}, H_{I}, H_{V}, H_{T}$} for global triplet, instrument, verb, and target features respectively. $H_{IVT}$ is drawn from a bottleneck layer, connected to an early layer of the ResNet-18 backbone. $H_{I}$ is taken from the WSL's output, and the last two are taken from the CAGAM.

Inside the MHMA, $H_{IVT}$ is processed by a self-attention head. A projection function maps it to three separate vectors - query, key, and value. A scaled dot product then transforms them into one refined feature. $H_{I}$ is fed to a cross-attention head: the key and value are obtained by projecting $H_{I}$, but the query is the same one used in $H_{IVT}$'s self-attention head. $H_{V}$, $H_{T}$'s cross-attention heads operate in the same manner as $H_{I}$'s.

The resulting four refined features are concatenated; two convolutions and an AddNorm operation complete the MHMA, which outputs a refined version of $H_{IVT}$. Inside Rendezvous, a stack of 8 MHMAs is employed; the output of this stack is used to make the final prediction on the triplet.

\subsection{Competing methods}

\team{Team 2Ai}
{Surgical video analysis using an ensemble of multi-task recurrent convolutional networks}
{Team 2Ai proposed a solution (version 1) for this challenge consisting of an ensemble of multi-task recurrent convolutional networks. Each model architecture consists of a multi-task recurrent convolutional network with four heads, where each branch targets one of four tasks, namely surgical instrument detection, verb recognition, phase identification, and target recognition. To extract generic visual features, they used a shared feature extractor for all four branches. Specifically, for each of the instrument detection and target recognition branches, a fully connected layer is connected to the backbone to compute the signals for both tasks. Here, a sigmoid activation layer is applied to produce the final predictions. In the branches for surgical action and phase recognition, long short-term memory (LSTM) units \citep{hochreiter1997long} are connected to the backbone to leverage the temporal context of the current frame. A sigmoid and softmax layers are added to the end of the last LSTM units in the action and phase recognition branches, respectively. Finally, binary cross-entropy is used as a loss function for the surgical instrument detection, verb recognition, and target recognition tasks, and cross-entropy is used for phase identification. The different networks, each with a different feature extraction backbone, are individually optimized and trained. Afterward, majority-vote ensemble is applied to combine the predictions resulting from the different networks. As a final step, a temporal smoothing technique is applied to each task to avoid temporally incoherent results. To meet the challenge output requirements, the individual signals are finally represented as surgical actions triplets. 2Ai version 2 is submitted post-challenge to correct the error in the output format of the initial model by changing the final output mapping from binary to probability scores. In this revised version, the ensemble is performed using an average of probabilities outputs from all the networks.}

\team{Team ANL-Triplet}
{Exploiting temporal information for triplet recognition}
{Team ANL-Triplet used three ResNet-18 backbones \citep{he2016deep} for instrument, verb and target prediction. In a multi-task setup, these component predictions are concatenated before using a single convolutional layer and a fully connected layer to recognize the action triplets represented in the frame. To improve verb and target recognition performance, these models are initialized using the learned instrument model weight. Additionally, an independent ResNet-34 is trained to incorporate temporal information by predicting action triplets directly from features extracted from the current frame and 7 previous frames. In both triplet recognition approaches, a smoothing factor of $0.2, 0.2^2$ is used for predictions that predicted two and one components correctly, respectively, reducing the penalty for semantically closer predictions. The final prediction is made by taking the average of the two triplet predictions.}

\team{Team Band of Broeders}
{YOLOv5 for surgical action triplet detection}
{The Band of Broeders utilized manually generated bounding box annotations of the surgical triplets to train a YOLOv5 based object detector for action triplet prediction. They chose the YOLOv5 \citep{Jocher_YOLOv5_by_Ultralytics_2020} for its state-of-the-art performance on various object detection datasets. Their network\footnote{non-competing method as it is trained on private annotations} consists of three stages: (1) the backbone CSP-DenseNet \citep{wang2020cspnet}, which is used for its gradient efficiency, (2) the neck Path Aggregation Network (PA-Net) \citep{liu2018path}, which is used for multi-scale feature extraction, and (3) the head which produces the final bounding box predictions. These components allow for the building of neural networks with large receptive fields and a multi-scale view of the frame enabling the detection of objects of various sizes.}

\team{Team CAMP}
{EndoVisNet: Phase-guided temporal endoscopic action triplet classification}
{Team CAMP's methodology is based on the idea that the presence of an action in a given frame implies that only a smaller set of actions could occur in the immediate frames that follow. To leverage this temporal property, visual features from both the input frame and its $t$ preceding frames are extracted using the SlowFast \citep{feichtenhofer2019slowfast} network to ensure that the extracted features are relevant and independent of the movement speed in an input video. After extraction, the temporal features are pooled into 1 dimension to make a final prediction. Feature classification is modeled using a similar approach to Tripnet \citep{nwoye2020recognition}, leveraging a multi-task learning approach, but with an extra branch for the phase detection, which is trained jointly on labels extracted from the Cholec80 dataset \citep{twinanda2016endonet}. The final action triplet classification scores are computed as a learned linear combination of the instrument, verb, and target prediction scores.}

\begin{figure}[h]
\centering
    \includegraphics[width=1.0\columnwidth]{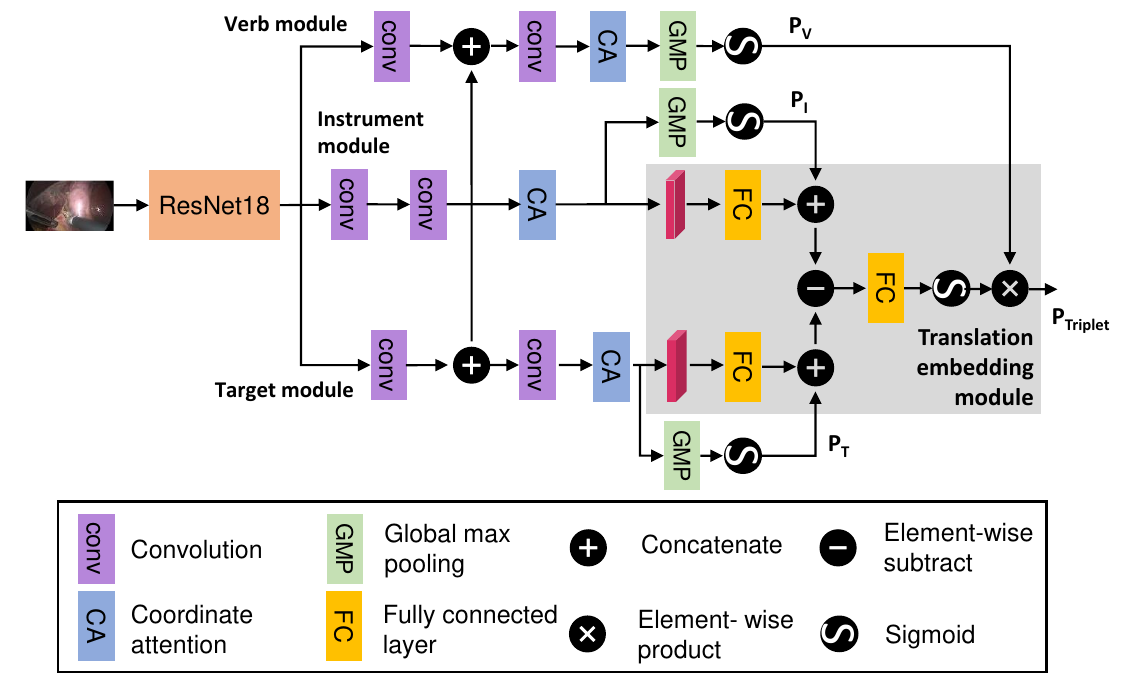} 
    \caption{Overview of the Casia Robotics submission.}
    \label{fig:teams_casia_robotics}
\end{figure}

\team{Team Casia Robotics}
{Triplet translation embedding network with coordinate attention}
{Team Casia Robotics' submission (see Fig. \ref{fig:teams_casia_robotics}) builds on the work \citep{nwoye2020recognition} and replaces the introduced 3D interaction space with a translation embedding module, following \cite{zhang2017visual}. Here, the translation embedding module couples the instrument, target, and verb features to produce the final triplet predictions. Specifically, they used a ResNet-18 \citep{he2016deep} backbone for extracting spatial features followed by three sub-networks with convolutional layers and fully-connected layers that are used to predict instruments, verbs, and targets, respectively. To model the complex relationship between triplets, they modeled projection matrices, $W_i$ and $W_t$, that translate learned representations from the feature space to a shared relation space. The interaction relation between the three learned component feature vectors $f_t$ (target), $f_i$ (instrument/tool), $f_v$ (verb) is then represented as:
$$
(W_tf_t - W_if_i).f_v.
$$
All invalid component combinations are then masked out to only generate predictions for the 100 relevant triplet classes. Further, a coordinate attention mechanism \citep{hou2021coordinate} is used to incorporate positional information into each sub-network to more accurately locate each component of the triplet.}

\team{Team Ceaiik} 
{Spatio-temporal learning of action triplets in surgical videos} 
{Team Ceaiik's model architecture design is based on the triplet classification as a composition of three individual tasks namely instrument, verb, and target classifications \citep{nwoye2020recognition}. In the first stage of training, the model utilizes a ResNet-50 \citep{he2016deep} to extract spatial features based on the results of \cite{mishra2017learning,mondal2019multitask} and employs 3 classification heads to predict instrument, verb, and target class probabilities. A final classification head is employed to utilize the previously computed component probabilities, which are aggregated to make a triplet prediction. Further, in the second stage of training, an effort is made to incorporate certain temporal properties of action triplets in their model design. Keeping the learned ResNet-50 weights frozen from the first stage of training, the model extracts visual features which are passed through an LSTM module \citep{hochreiter1997long} before instrument, verb, target, and triplet prediction are performed using a similar approach to the first stage.}

\team{Team CITI SJTU}
{Action triplet recognition via convolutional LSTMs and multi-task learning} 
{Team CITI SJTU focused their method on modeling the sub-components and temporal coherence when predicting action triplets. Their multi-task deep learning network includes four branches with one main triplet branch and three auxiliary branches generating the recognition results for instruments, verbs, and targets, respectively. All the branches share the same ResNet-50 \citep{he2016deep} network for feature extraction, followed by two convolutional Long Short Term Memory (ConvLSTM) layers \citep{shi2015convolutional} for modeling spatial-temporal relationships. After training, they used the triplet prediction branch to obtain the final triplet prediction for inference. The three auxiliary branches allow them to add fine-grained information for training to significantly boost triplet recognition performance.}

\team{Team Digital Surgery}
{TE-TAR: Temporal ensemble triplet action recognition } 
{Team Digital Surgery's proposed model, TE-TAR, consists of an ensemble of encoders, an LSTM model \citep{hochreiter1997long}, and a classification layer. The ensemble of encoders is composed of four HRNet32 backbones \citep{wang2020deep} and a classification head that efficiently combines multi-scale information. Each ensemble is trained with a different subset of data for learning more diverse features. For each image, four features are generated in total (i.e., one per encoder), which are combined by summation. In addition, to allow the model to estimate the action of the instrument (verb) and the anatomy where the instrument is applied (target), they proposed the use of temporal information, using a sliding window approach with a window length of 10 frames. Features are encoded by the ensemble for all frames in the window and then fed to an LSTM. As they formulated the action triplet task as a classification problem, the final triplets are estimated by feeding the aggregated features to a linear classification layer.}

\team{Team Lsgroup} 
{Feature fusion and weak locational information calculating in triplet classification multi-task} 
{Team Lsgroup submitted a multi-task learning network with four sub-networks that are used to classify each component of the action triplet and the triplet itself. Their model is based on two fundamental assumptions: (1) The type and location of surgical instruments are critical factors for determining the verb component of the triplet \citep{nwoye2020recognition}, and (2) it is important to combine the feature of the instrument, verb, and target \citep{nwoye2020recognition,jin2021temporal}. Therefore, their approach is composed of a CNN backbone (feature extraction), weak instrument localization, component feature fusion sub-network, and triplet-possibility mapping using weighted averaging. Firstly, a ResNet-18 backbone is trained to jointly learn to perform instrument, verb, and target prediction using three classifier heads. To focus the predictions on the relevant portions of the image, weak instrument localization features \citep{nwoye2019weakly} are concatenated to the learned features for verb and target heads before performing the final classification. The predicted probability vectors for each of the three tasks are then concatenated before a single fully connected layer is used to map this vector to the final triplet prediction vector.}

\begin{figure}[]
\centering
    \includegraphics[width=1.0\columnwidth]{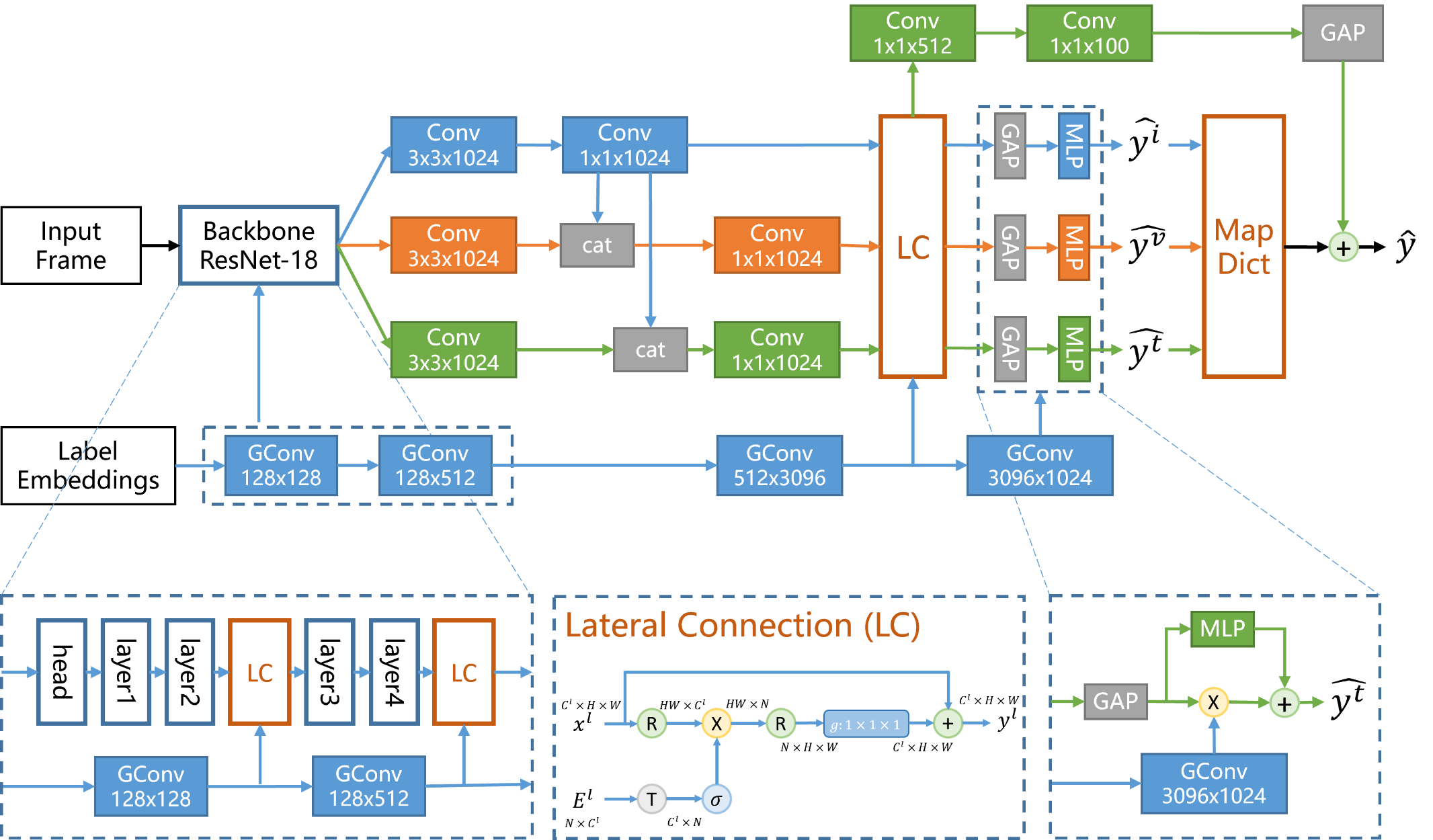} 
    \caption{Overview of the HFUT-MedIA submission.}
    \label{fig:teams_hfut_media}
\end{figure}

\team{Team HFUT-MedIA} 
{COEMNet: Correlation embedded multi-task network} 
{Team HFUT-MedIA participated with a correlation embedded multi-task network, named COEMNet (see Fig. \ref{fig:teams_hfut_media}). First, a multi-task learning network is trained for instrument, verb, and target recognition tasks. The learned features for instrument prediction are leveraged for better recognition of verbs and targets. Secondly, the correlation between all classes is modeled as a graph and used the statistical co-occurrence frequencies as the adjacency matrix. A multi-layer graph convolutional network (GCN) is deeply integrated into their end-to-end network to efficiently learn the label classifiers and embed correlation information into feature representation {\citep{wang2020multi}}. Since associating the predicted instruments, verbs, and targets to form the right triplet predictions is complex when there are multiple triplet annotations in a single frame, a triplet adjustment task is added to the proposed network to minimize the association errors. The learned adjustment factors are applied to the original triplet prediction for more accurate recognition.}

\team{Team HFUT-NUS} 
{Multi-task learning-based surgical interaction triplet recognition} 
{Team HFUT-NUS's method utilizes a ResNet-18 \citep{he2016deep} backbone to extract features from endoscopic images followed by three parallel fully connected layers that are used to predict instrument, verb, and target, respectively. To further improve the precision of surgical interaction triplet detection, an attention mechanism is adopted to weigh the importance of instrument, verb, and target components. Finally, the weighted instrument vector, verb vector, and target vector are spliced together and fed into a fully connected layer with a sigmoid activation to predict the final surgical action triplet. }

\team{Team J\normalfont{\&}M} 
{Surgical action triplet recognition with Efficient-MSTCN} 
{Team J\&M's submission is a 2-stage network that is trained to first extract relevant spatial features and then utilize the temporal context of each frame to make action triplet predictions for a frame. In their proposed model, EfficientNetV2-M \citep{tan2021efficientnetv2} is used as the feature extractor. It is trained to make predictions using a single frame. Afterward, the weights are frozen and a Multi-Stage Temporal Convolutional Network (MS-TCN) \citep{farha2019ms} is employed to refine the model predictions.}

\team{Team Med Recognizer} 
{Surgical action triplet recognition via temporal memory relation gradient network} 
{Team Med Recognizer's competing model is a temporal memory relation gradient network, in which a stem module first extracts spatial features from each frame, and then splits into three branches, with different temporal supportive information integrated to represent the action triplet. The temporal lengths are set as 10, 5, and 5 for ‘verb’, ‘instrument’, and ‘target’, respectively, given that different tasks require different amounts of temporal context. The spatial-temporal feature extraction model is developed based on \cite{jin2021temporal}, which is originally designed for phase recognition, to leverage the long-range and multi-scale temporal patterns in the video. The features are then fed into the classifiers to generate the prediction probabilities of the three tasks. The obtained probabilities from three branches are then integrated to produce the final one. Post-processing methods are then applied to account for the label imbalance, where higher weights are assigned to the classes with fewer data samples. The temporal information is also further utilized to weight the predicted probabilities of the previous frames when producing the results of the current frame. Dropout is used at both the training and the testing time.}

\team{Team MMLAB} 
{Temporal triplet net for triplet presence detection in surgical videos} 
{Team MMLAB proposed the Temporal Triplet Net (TTN) which consists of DenseNet \citep{huang2017densely} and a Graph Convolutional Network (GCN) that utilizes both spatial and temporal features for the recognition of action triplets in surgical videos. The DenseNet is used as an image classification model to extract spatial features for each labeled frame for an input video. To utilize the temporal information, the GCN is incorporated to capture the temporal relationships among the continuous frames of a video sequence using the features extracted for each frame. Specifically, the representation for each frame is regarded as the node of the graph, and the similarity between each pair of nodes is regarded as the edge of the graph.}

\team{Team NCT-TSO} 
{Multi-task learning framework for action triplet recognition} 
{Team NCT-TSO's method is similar to  \cite{nwoye2020recognition} designed for the tasks of verb, target, and triplet recognition. In their proposed method, a separate model is trained for instrument recognition using the Cholec80 dataset \citep{twinanda2016endonet}. The instrument recognition model is based on a convolutional neural network (CNN) which uses the ResNet-50 \citep{he2016deep} as its backbone and spatial pooling to learn class-specific feature maps of the instruments in a weakly supervised manner \citep{durand2017wildcat}. These instrument maps are subsequently fed to the verb and target paths of the triplet recognition network. The verb and target paths share the same ResNet-50 model as their backbone, followed by two convolutional layers for each path. The verb and target paths are also trained to learn class-specific feature maps and use wildcat spatial pooling \citep{durand2017wildcat} on these maps for the prediction of labels. The instrument logits from the pre-trained instrument model, along with the verb and target logits, are subsequently used to learn the triplets using a 3D interaction space as proposed in the work of \cite{nwoye2020recognition}. }

\begin{figure}[]
\centering
    \includegraphics[width=1.0\columnwidth]{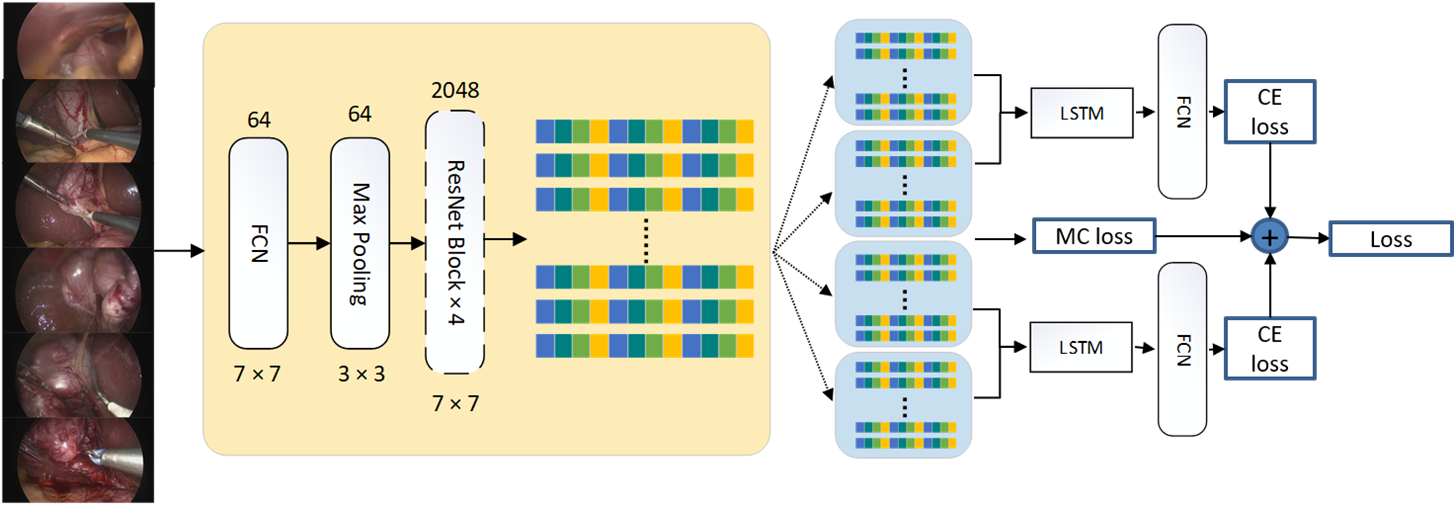} 
    \caption{Overview of the SIAT CAMI submission.}
    \label{fig:teams_siat_cami}
\end{figure}

\team{Team SIAT CAMI} 
{Multi-task mutual Channel Recurrent Net for Fine-grained Surgical Triplet Recognition} 
{Team SIAT CAMI used MT-MCLNet (see Fig. \ref{fig:teams_siat_cami}), a multi-task surgical triplet recognition network with multi-label mutual channel loss \citep{chang2020devil} to extract local fine-grained features and aggregate temporal information on verb and triplet branch. When a sequence of video images is fed into the network, the backbone ResNet-50 module \citep{he2016deep} first extracts a global 2048-dimension spatial feature for each image in the sequence. To extend the global features to each task, different $1\times 1$ convolutions are applied to generate 2040-dimension features for instrument and target branches, 2000 for verb and triplet. Then, the 2040/2000-dimension spatial features are fed into two LSTM modules to capture 512-dimension temporal motion information. Then, 4 fully connected layers are utilized to produce the final classification output. The overall loss function is a weighted sum of standard cross-entropy loss and a mutual channel loss, which is intended to encourage learning a diverse range of discriminative features.}

\team{Team SJTU-IMR} 
{Tracking surgical actions with transformers and action label-guided fine-grained information aggregation} 
{Although each triplet is unique, different triplets may share certain elements. Team SJTU-IMR hypothesized that this observation is critical to modeling instrument-tissue interactions as it can establish a link between different actions with shared elements. To this end, they presented a two-stage transformer-based learning framework to solve the surgical action tracking problem. Specifically, the features learned from the first stage, which conducts frame-level action recognition via a Swin-S network architecture \citep{liu2021swin}, are taken as the input to the second stage, which performs sequence-level action recognition via masked transformers \citep{vaswani2017attention}. At both stages, the same multi-task learning strategy of combining coarse-grained action recognition with fine-grained information aggregation is employed. Their fine-grained information aggregation is guided by coarse-grained action labels so that the networks can put more emphasis on features modeling instrument-tissue interactions.}

\team{Team SK} 
{Action triplet recognition with weakly-supervised attention of surgical instruments}
{Team SK followed an instrument-centric approach to action triplet recognition by first performing instrument recognition that is then used to condition the verb, target, and consequently, triplet recognition tasks. Using the first 4 convolutional layers of a ResNet-18 \citep{he2016deep} as their backbone, the first sub-network generates attention maps using 2 convolutional layers (layer 5 of ResNet-18 + 1x1 convolution) to localize the position of the instruments in the image. These attention maps are implicitly learned through weak supervision with instrument labels. A second sub-network then uses a convolutional layer to predict a fixed number of channels corresponding to each instrument which is multiplied by the previously learned attention maps to appropriately weigh different parts of the image based on instrument location. Finally, a 1x1 convolution, global average pooling, and softmax operation are performed to obtain the triplet probabilities.}

\team{Team Trequartista} 
{Phase-aware multitasking surgical action recognition model with adjustment for low-data triplet classes}
{Team Trequartista's entry used multi-task learning on 5 tasks, triplet, instrument, verb, target, and phase prediction. The phase prediction ground truth made use of Cholec80 annotations \citep{twinanda2016endonet}. Using a ResNet-18 and ResNet-34 \citep{he2016deep} as their primary models, their work focuses on finding the optimal hyperparameters and appropriately post-processing their predictions specifically for the challenge metric (mAP). Following an observation that predicting under-represented classes has a strongly negative effect on the overall metric, the triplet predictions probability for the 63 least represented triplets are adjusted to compensate for this effect based on the following heuristic:
\begin{equation}
\label{eqn:heu}
    \resizebox{.85\hsize}{!}{
    {\bf P}(Triplet) = {\bf P}(Instrument) * 0.03{\bf P}(Verb) + 0.97{\bf P}(Target).
    }
\end{equation}
}

\subsection{Theoretical comparative analysis}
Presented methods are broadly classified into 5 categories:

\method{Multi-task learning}
{Rank 1-5, 7-8, 10-12, 14-19}
{Multi-task learning methods aim to improve triplet recognition performance by learning a shared representational space by training a model to perform multiple related tasks. Unsurprisingly, given the availability of additional annotated information, all but 3 teams employed multi-task learning in their model design. The most common formulation is to learn to predict the triplet components, instrument-verb-target, in addition to the triplet itself. Interestingly, this formulation has come in two flavors: 1) Predicting the triplet as an explicitly modeled association of the 3 predicted components (2Ai, ANL Triplet, CAMP, HFUT-MedIA, HFUT-NUS, lsgroup, Med Recognizer, NCT-TSO, SJTU-IMR, and 2) Predicting the triplets and 3 components using a shared backbone that implicitly learns useful features from the other tasks for triplet recognition (CITI SJTU, Trequartista, SIAT-CAMI). Given that surgical phases \citep{twinanda2016endonet} are defined by and consequently signal the occurrence of certain actions, surgical phase recognition is highly related to the task of action triplet recognition. Three teams (Trequartista, 2Ai, CAMP)  leverage this fact by incorporating spatial phase annotations for the challenge training videos that are publicly available in the Cholec80 dataset \citep{twinanda2016endonet}. All three methods do so in multi-task learning setups, with phase and triplet component recognition included as relevant tasks to boost triplet recognition performance. Finally, the Band of Broeders entry treated the triplet recognition tasks as an object detection task, positing that localization information is critical to effectively recognizing the action triplets represented in an image. They manually generated bounding boxes and instrument labels, which are then assigned to their corresponding triplets, to facilitate model training by learning to both detect and localize triplets.
 }
 
\method{Temporal modeling}
{Rank 2, 4-6, 8-11, 18-19}
{While surgical action triplet recognition can be done on single frames, temporal models processing information from several frames at a time are interesting solutions to investigate: object permanence, motion, and surgical workflow are indeed temporal concepts that can inform action recognition. The many teams (11 out of 19) choosing this direction proposed a diverse range of temporal modeling methods, which can be described according to three main traits. The first is the choice of temporal architecture. Some methods used non-trainable operations to aggregate information from static image models across multiple frames (ANL-Triplet, LSGroup, CAMP, MMLAB); most entrants, however, resort to sequence models running on features extracted by CNN. Among them, LSTM-based recurrent neural networks are a clear trend, with various forms appearing in 6 submissions (2AI, CEAIIK, Digital Surgery, CITI-SJTU, Med Recognizer, SIAT-CAMI). Team CITI-SJTU in particular used a pair of ConvLSTMs. Besides LSTMs, other more recent models appeared as well: TCNs (J\&M, Med Recognizer), and Transformers (SJTU-IMR).

The second notable trait is the model's temporal range. In some methods, this range is very short: 2 frames for team LSGroup, 4 frames for team CEAIIK, 5 frames for team ANL-Triplet, and team MMLAB. Longer periods are featured in submissions by team 2AI (10 frames), team Digital Surgery (10 frames), team CITI-SJTU (16 frames), and team CAMP (32 frames); team Med Recognizer used "short video clips" for the SV-RCNet, 10-frame clips for the verb memory bank, and 5-frame clips for the instrument memory bank. Team SJTU-IMR feed their model in chunks of 200 frames, which can cover large workflow sections. One team, J\&M, used concatenated features from all video frames, making the entire history available to the model at any given timestep.

The third trait distinguishing temporal methods is end-to-end training. Some methods feature temporal layers that are trained separately (CEAIIK, Digital Surgery, J\&M, Med Recognizer, SJTU-IMR), while others (2AI, ANL-Triplet, CITI-SJTU, SIAT-CAMI, LSGroup, CAMP) trained all parts simultaneously.
}

\method{Attention mechanism/transformer}
{Rank 3,5,9,10,12,14}
{Attention mechanisms are methods designed to modulate a model's input or internal representation, to highlight parts that are important for making predictions. The use of attention can greatly improve surgical action triplet recognition, as shown by \cite{nwoye2021rendezvous} in the two approaches of Attention Tripnet and Rendezvous - these two models reappear in this challenge as baselines. All methods from the challenge featured in this category used some form of spatial attention since areas surrounding tooltips are particularly informative. A few of these methods explicitly used instrument maps as the source of attention (Attention Tripnet, RDV, LSGroup, SK). The other forms of spatial attention featured are team SIAT-CAMI's "channel-wise attention", team Casia Robotics's "coordinate attention" \citep{qibin2021coordinate} and team SJTU-IMR's Swin-S vision Transformer. Two entries used additional attention components of different types: semantic, in the RDV baseline's multi-head of mixed attention; and temporal, in the masked Transformer appearing in the second stage of team SJTU-IMR's method.
}

\method{Graph convolutional networks}
{Rank 3, 13}
{Two teams, HFUT-MedIA and MMLAB, made use of GCN and CNN to better recognize action triplets represented in a given image, using two different strategies. Team MMLab used a GCN primarily to leverage temporal relationships using spatial features extracted using a CNN in the first stage of training. Here, the extracted features for each frame are treated as a graph node and the similarity between each pair of nodes features an edge between two nodes. Team HFUT-MedIA, in contrast, does not base its graph on CNN-extracted spatial features but rather used it to effectively incorporate triplet co-occurrence distribution statistics into various stages of a discriminative CNN to predict action triplet probabilities. }

\method{Ensemble models}
{Rank 1, 6, 16}
{Ensembling is a commonly used strategy for limiting noisy predictions, by combining outputs from independently trained models. Three teams employed this type of approach, with distinct variations on the ensembling concept. The choice of operation for merging outputs varies between teams: summation (Digital Surgery), averaging (2AI version 2), majority vote (2AI version 1), ad-hoc heuristics (Trequartista). Early or late fusion is another differentiating trait: teams Trequartista and 2AI merged final probabilities, while team Digital Surgery combined features from various feature extractors before applying an LSTM model. Finally, various ensemble sizes and architectures are used: 3 ResNets (Trequartista), 4 HRNets (Digital Surgery), and an ensemble of 6 CNNs (team 2AI).}

\subsection{Ensemble predictions}
For optimal performance on the dataset and as a summary of the challenge benchmark study, we ensemble the predictions of 7 top models (with triplet recognition AP above 30.0\%): Trequartista, HFUT-MedIA, SIAT-CAMI, RDV, ANL-Triplet, CITI-SJTU, and Digital Surgery. 
This helps minimize errors due to noise, bias, and variance while improving the stability, reliability, and accuracy of predictions.
In this work, we experiment with 6 ensemble methods.

\begin{figure}[]
\centering
    \includegraphics[width=1.0\columnwidth]{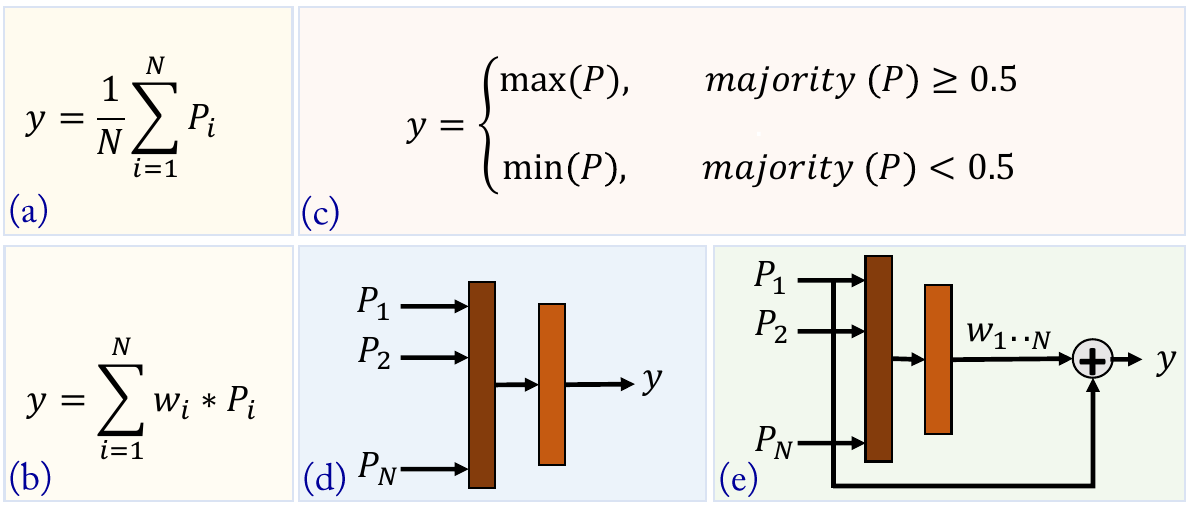} 
    \caption{Ensemble method for the model predictions: (a) averaging, (b) weighted averaging, (c) majority soft voting, (d) deep ensemble, and (e) deep weighted ensemble.}
    \label{method:ensmble}
\end{figure}

As shown in Fig. \ref{method:ensmble}, we start with simple {\bf averaging} of the probability scores of the different models (Fig. \ref{method:ensmble}a). This is extended to {\bf weighted averaging} (Fig. \ref{method:ensmble}b) where the weights are computed as the performance ratio of each model against the others. We also perform {\bf soft voting} (Fig. \ref{method:ensmble}c) between the maximum (presence) and minimum (absence) probability scores per class at a threshold of 0.5.
Our first trainable method uses a two-layer network to learn a {\bf deep ensemble} (Fig. \ref{method:ensmble}d) of the challenge networks predictions.
Lastly, we focus on learning the model averaging weights and in turn use this for a {\bf deep weighted ensemble} (Fig. \ref{method:ensmble}e). In this case, we learn two types of weights: (1) a vector of $N$ weights for the $N$ models, and (2) a matrix of $N\times C$ weights for per class ($C$) weights of $N$ models. The latter option, which is a {\bf deep per-class weighted ensemble}, is designed to utilize the strength of each model in recognizing the different categories of the triplets.
All the deep ensemble models are trained on triplets probability ($Y_{IVT}$) predictions of the selected models on the training dataset.

\subsection{Implementation details}
The implementation details of all submissions as well as the baselines are summarized in Table \ref{tab:implementation}.

\begin{table*}[htp]
    \caption{Methodological and implementation details.} 
    \label{tab:implementation}
    \resizebox{1.0\textwidth}{!}{%
        \includegraphics[width=1.0\linewidth,height=0.98\textheight,]{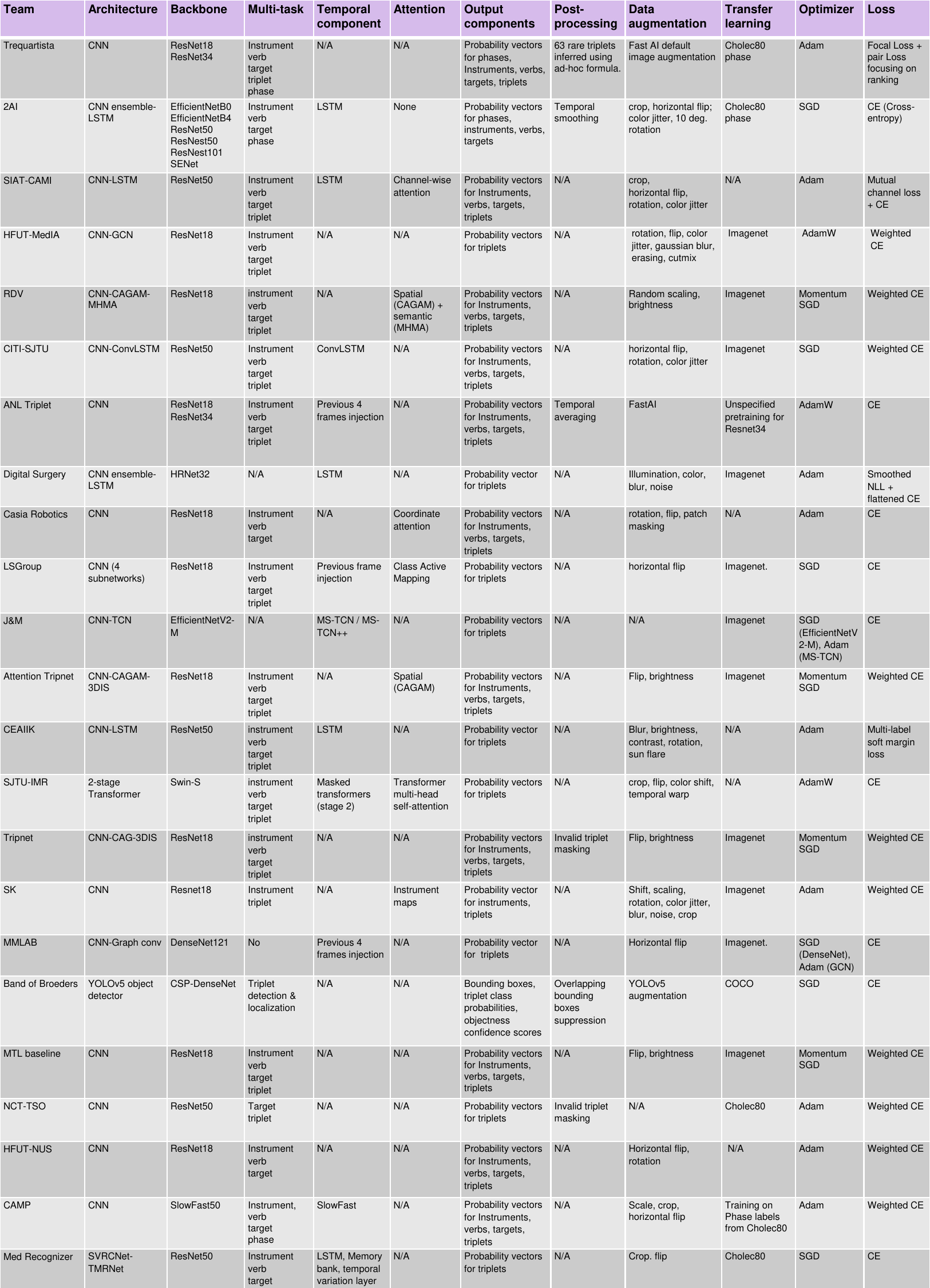}
    }
\end{table*}

%% file: main/06-evaluation.tex
\section{Evaluation }

\subsection{Metrics}
To evaluate the performance of the presented models on surgical action triplet recognition, we use the average precision (AP) metric, computed as the area under the precision-recall curve; this is recommended as the standard practice for the CholecT50 dataset.
Using AP metrics, we access the models' capacity for predicting the correct triplet components: AP$_I$, AP$_V$, AP$_T$, and the correct triplet associations: AP$_IV$, AP$_IT$, AP$_IVT$. The AP$_IVT$ evaluates the complete instrument-verb-target combination and hence serves as the primary metric in the challenge.
We also analyze the quality of the model predictions by computing the top$K$ performance scores where $K\in [5,10,15,20]$ and the average over top$K$@[5:20] at intervals of 5 steps. All the evaluation scores are computed using the \path{ivtmetrics} library\footnote{\url{https://pypi.org/project/ivtmetrics}} \citep{nwoye2022data}.

\subsection{Evaluation protocol}
We maintain an online inference strategy to represent the real-time usage of developed methods in the OR.
For uniform evaluation of all models - including those not multitasking the triplet components - we collect only the probability scores for triplet classes ($Y_{IVT}$) per frame as the model output. The individual component predictions are filtered from $Y_{IVT}$ following the disentanglement function proposed in \cite{nwoye2021rendezvous}. 
The video-specific AP scores are computed per category in a test video, then averaged categorically across all test videos. We obtain the mean AP by averaging the category APs.

For method ranking, the challenge evaluation is performed on 94 valid triplet classes excluding the 6 null classes in the dataset. The null classes, which include \triplet{grasper, null-verb, null-target}, \triplet{bipolar, null-verb, null-target}, \triplet{hook, null-verb, null-target}, \triplet{scissors, null-verb, null-target}, \triplet{clipper, null-verb, null-target}, and \triplet{irrigator, null-verb, null-target}, are possible only when an instrument is idle or performing an action that is not part of the 100 considered triplet classes. A total absence of a triplet, \triplet{null-instrument, null-verb, null-target}, is not an explicit class in a multi-label classification problem.

%% file: main/07-results.tex
\section{Results and discussion }
\begin{table*}[thbp]
    \centering
    \caption{Performance summary of the presented methods across all task divisions}
    \label{tab:results:quantitative:summary}
    \setlength{\tabcolsep}{3pt}
    \resizebox{0.9\textwidth}{!}{%
    \begin{tabular}{@{}lrcccrcccrr@{}}
    \toprule
         & \phantom{abc} &
        \multicolumn{3}{c}{Component detection} & \phantom{abc} &
        \multicolumn{3}{c}{Triplet association} & \phantom{abc} &
        \multirow{2}{*}{\makecell[c]{Challenge\\ranking}} \\ 
        \cmidrule{3-5} 
        \cmidrule{7-9} 
        Team && $AP_{I}$ & $AP_{V}$ & $AP_{T}$ && $AP_{IV}$ & $AP_{IT}$ & $AP_{IVT}$ &&\\ 
        \midrule
        Trequartista && 79.9 & \textbf{52.9} & \textbf{46.4} && \underline{39.0} & 41.9 & \textbf{38.1} && 1\\         
        2AI: Version 2 $^\dag$ && 79.8 & 50.1 & 42.8 && 35.2 & \underline{42.4} & \underline{36.9} && -\\ 
        SIAT CAMI && \textbf{82.1} & \underline{51.5} & \underline{45.5} && 37.1 & \textbf{43.1} & 35.8 && 2\\ 
        HFUT-MedIA && 77.1 & 46.7 & 37.8 && 33.1 & 35.9 & 32.9 && 3\\ 
        RDV (CAMMA) $^\ddag$ && 77.5 & 47.5 & 37.7 && \textbf{39.4} & 39.6 & 32.7 && -\\ 
        CITI SJTU && 67.8 & 37.1 & 34.8 && 29.9 & 33.0 & 32.0 && 4\\ 
        ANL Triplet && 73.6 & 47.3 & 40.5 && 32.6 & 37.1 & 31.9 && 5\\ 
        Digital Surgery && \underline{80.8} & 50.0 & 41.1 && 35.1 & 35.7 & 31.7 && 6\\ 
        Casia Robotics && 72.6 & 43.9 & 31.2 && 30.7 & 30.6 & 26.7 && 7\\ 
        Lsgroup && 73.8 & 44.3 & 34.9 && 31.4 & 31.9 & 26.3 && 8\\ 
        J\&M && 69.4 & 46.7 & 39.2 && 28.9 & 28.8 & 25.6 && 9\\ 
        Attention-Tripnet (CAMMA) $^\ddag$ && 77.1 & 43.4 & 30.0 && 32.3 & 29.7 & 25.5 && -\\ 
        Ceaiik && 68.9 & 40.5 & 30.9 && 27.5 & 28.4 & 25.2 && 10\\ 
        SJTU-IMR && 72.6 & 42.5 & 34.1 && 29.2 & 26.4 & 24.8 && 11\\ 
        Tripnet (CAMMA) $^\ddag$ && 74.6 & 42.9 & 32.2 && 27.0 & 28.0 & 23.4 && -\\ 
        SK && 52.6 & 30.4 & 20.2 && 25.8 & 21.0 & 18.4 && 12\\ 
        MMLAB && 50.1 & 31.8 & 31.6 && 20.6 & 22.1 & 18.1 && 13\\ 
        Band of Broeders $^\P$ && 63.4 & 35.2 & 26.2 && 19.7 & 18.6 & 16.0 && -\\ 
        MTL baseline (CAMMA) $^\ddag$ && 48.6 & 27.8 & 19.8 && 18.5 & 15.5 & 13.7 && -\\ 
        NCT-TSO && 27.3 & 15.7 & 12.3 && 13.6 & 11.7 & 10.4 && 14\\ 
        2AI: Version 1 && 46.2 & 24.4 & 20.5 && 13.3 & 12.3 & 10.0 && 15\\ 
        HFUT-NUS && 34.1 & 20.2 & 13.5 && 16.0 & 11.2 & 09.8 && 16\\ 
        CAMP && 30.4 & 19.5 & 11.8 && 13.2 & 09.7 & 09.3 && 17\\ 
        Med Recognizer && 20.6 & 12.8 & 10.1 && 07.0 & 04.5 & 04.2  && 18\\
        \midrule
        Mean ± standard deviation (stdev) && 62.5±18.9 & 37.7±12.41 & 30.2±11.0 && 26.3±8.9 & 26.5±11.3 & 23.3±9.9  &&\\
        \bottomrule 
        \multicolumn{11}{l}{\makecell[l]{\footnotesize{
		{\bf bold} = best score and \underline{underlined} = second best. Not eligible for award: $\dag$ post-challenge submission, $\ddag$ organizers' baselines, $\P$ used non-public third-party dataset. }}}\\
    \end{tabular}
    }
\end{table*}

In this section, we provide a quantitative and qualitative overview of all methods featured in the challenge, including baselines and models evaluated post-challenge.
In total there are 24 models: 4 are baseline models from the challenge organizers, 19 are competing models and 1 is a post-challenge submission.
{\red The baseline models provide  lower bound performances for monitoring and assessing the improvement and effectiveness of newer implementations.}
The analysis of their results builds a foundation for the CholecT50 dataset, establishing it as a validated reference benchmark.

\subsection{Summary of the quantitative results}
For a concise overview, we first summarize the AP scores on both the component detection and triplet association in Table \ref{tab:results:quantitative:summary}.
On the instrument component, half of the presented models achieved an AP score higher than 70\% with the highest score of 82.1\% by team SIAT-CAMI and the average performance (with standard deviation) of 62.5±18.9\%. 
These results suggest the tremendous progress made in the use of deep learning models for surgical instrument recognition in laparoscopic videos, as 18 out of the 24 presented methods recognized the instruments at a performance higher than 50\%.  
The verb recognition peaked at 52.9\% AP with only four teams achieving a higher than 50\% score. It is observed that these four teams either leveraged temporal information or exploited phase labels, which is also a temporal task. This suggests a strong correlation between surgical phases and actions and the quality of temporal feature modeling in tracking activity workflow.
Improving the verb performance is a promising direction for future work likely by exploring a better-conditioned range of temporal dependencies attuned to triplet timings. The bulk of the AP scores from the competing teams falls within 30-40\% with a mean of 37.7±12.1\%.

Target, being the most difficult component, was recognized at a maximum AP of 46.4\% by team Trequartista. The low performance on this sub-task can be attributed to the \textit{instrument-centric} nature of the underlying target which makes its recognition very challenging. The mean performance at the challenge was 30.2±11.0\%. As evidenced by the submissions, tackling surgical target recognition still is not straightforward.

On the association part, an interesting observation is that top-performing models recognized the instrument-target (AP$_{IT}$) pair better than they recognized the instrument-verb (AP$_{IV}$) pair, despite the larger number of classes for the former. Their drop in performance from AP$_T$ to AP$_{IT}$ is much lower than from AP$_{V}$ to AP$_{IV}$. These deep learning models were more likely to recognize the correct operating instruments given an underlying target (a spatial relationship) than given their actions (a temporal relationship). The reverse is the case for lower-performing models. Here, it is easier to recognize instrument-verb pairs as most instruments perform specific actions and the verbs have a lesser number of classes than targets, and therefore their combinations (IV or IT). Many instruments can act on a wide range of targets - including unintended contact - making the problem more challenging for less advanced models. Overall these observations give a different perspective when interpreting the strength of the proposed models in understanding tool-tissue interactions. As shown in Table \ref{tab:results:quantitative:summary}, team SIAT-CAMI obtained the highest AP$_{IT}$ score while the baseline Rendezvous \cite{nwoye2021rendezvous} still retained the state-of-the-art (SOTA) performance on AP$_{IV}$.

The complete triplet recognition was best achieved at an AP of $38.1\%$ by team Trequartista topping the challenge leaderboard. The second was a model submitted post-challenge by 2AI as an improvement of their challenge competing method. Meanwhile, team SIAT-CAMI claimed the runner-up prize with an AP of 35.8\% while team HFUT-MedIA took the third-place prize with an AP of 32.9\%. Three other teams (CITI-SJTU, ANL Triplet, and Digital Surgery) and the baseline Rendezvous achieved an AP higher than 30\% which are promising performances for 100 class triplet recognition tasks. The mean performance for the triplet recognition recorded at the CholecTriplet 2021 challenge was 23.3±9.9\% suggesting room for improvement on this challenging task.

\subsection{Top$K$ accuracy on surgical action triplet recognition }
\begin{table}[]
    \centering
    \caption{Top K accuracy of the triplet predictions}
    \label{tab:results:quantitative:topN}
    \setlength{\tabcolsep}{3pt}
    \resizebox{\columnwidth}{!}{%
    \begin{tabular}{@{}lccccr@{}}
        \toprule
        Team &  Top 5 & Top 10 & Top 15 & Top 20 & Top \{5:20\}\\  
        \midrule
        RDV $^\ddag$  & \textbf{69.35} & \underline{84.38} & 89.93 & 93.24 & \textbf{84.23} \\ 
        Tripnet $^\ddag$   & 67.89 & 83.99 & 90.76 & \textbf{93.65} & \underline{84.07} \\ 
        HFUT-MedIA  & 65.05 & \textbf{85.35} & \underline{91.75} & \underline{93.59} & 83.94 \\ 
        Attention-Tripnet $^\ddag$  & 66.86 & 82.49 & \textbf{91.85} & 93.25 & 83.61 \\ 
        Trequartista   & \underline{68.50} & 82.40 & 88.24 & 92.29 & 82.86 \\ 
        Ceaiik   & 66.02 & 81.34 & 89.74 & 93.40 & 82.63 \\ 
        Digital-Surgery   & 65.97 & 81.56 & 88.78 & 92.92 & 82.31 \\ 
        HFUT-NUS   & 65.71 & 84.18 & 88.68 & 90.43 & 82.25 \\ 
        SIAT-CAMI  & 66.58 & 81.93 & 88.59 & 91.84 & 82.24 \\ 
        SJTU-IMR   & 66.50 & 81.88 & 84.19 & 84.89 & 79.37 \\ 
        ANL-Triplet   & 52.12 & 83.65 & 89.19 & 91.37 & 79.08 \\ 
        CITI-SJTU   & 54.95 & 78.61 & 88.96 & 92.41 & 78.73 \\ 
        SK &   48.08 & 79.46 & 90.41 & 92.12 & 77.52 \\ 
        2AI: Version 2 $^\dag$  & 64.87 & 76.17 & 82.54 & 86.26 & 77.46 \\ 
        Casia-Robotics   & 59.11 & 75.03 & 84.44 & 90.41 & 77.25 \\ 
        MMLAB &   60.53 & 76.57 & 82.72 & 86.67 & 76.62 \\ 
        Lsgroup  & 62.88 & 73.03 & 78.64 & 81.98 & 74.13 \\ 
        J\&M &   56.09 & 66.36 & 72.36 & 76.90 & 67.93 \\ 
        MTL-baseline $^\ddag$   & 45.59 & 52.45 & 56.65 & 59.77 & 53.62 \\ 
        Med-Recognizer   & 30.26 & 43.69 & 58.24 & 68.05 & 50.06 \\ 
        Band-of-Broeders $^\P$  & 39.86 & 40.34 & 41.14 & 48.27 & 42.40 \\ 
        2AI & 29.44 & 30.18 & 32.11 & 41.14 & 33.22 \\ 
        CAMP   & 08.73 & 17.91 & 21.25 & 25.71 & 18.40 \\ 
        NCT-TSO  & 04.88 & 11.78 & 16.59 & 21.93 & 13.80 \\
        \midrule
        \small{Mean ± stdev} & \small{53.1±18.3} & \small{67.5±22.3} & \small{73.9±23.3} & \small{77.9±22.0} & \small{68.1±21.2} \\
        \bottomrule 
        \multicolumn{6}{l}{\makecell[l]{\scriptsize{{\bf bold} = best score and \underline{underlined} = second best. Not eligible for award: $\ddag$ organizers' baselines,}\\ \scriptsize{$\dag$ post-challenge submission, $\P$ used non-public third-party dataset. }}}
    \end{tabular}
    }
\end{table}

Due to the large number of classes and the high semantic overlap in the triplet classes, we also evaluate the top$K$ of the presented models. This metric measures the ability of a model to predict the exact triplets within its top $K$ confidence scores. We analyze the top 5, 10, 15, 20, and average across these four thresholds, top$K$@[5:20] as shown in Table \ref{tab:results:quantitative:topN}. The obtained results describe the model's confidence in its predictions with the Rendezvous model obtaining a 69.35\% accuracy score as the best model when the top 5 confident predictions are considered. Similarly, HFUT-MedIA, Attention Tripnet, and Tripnet models produce the best results at top 10, 15, and 20 confidence scores respectively.

On average, the Rendezvous model can correctly recognize the triplets at a performance of 84.23\% when top confident predictions are taken into account. The average performance of all the presented models at the challenge is 53.1±18.3\%. On this metric, it is not surprising to see the challenge winner, Trequartista, scoring lower in top K accuracy because the model uses a mathematical operation to suppress the less confident predictions. These low confidence scores, when taken into account by the AP metric, lower the performance of other models. With top$K$ focusing only on top confident predictions, the models are not penalized by their less confident predictions. This accuracy metric is highly informative and more usable when thresholding predictions to binary values to obtain the unique IDs of the predicted triplets. It also suggests that most of the presented models would ordinarily obtain higher scores with fewer triplet classes, less class similarity, and less semantic overlap.

Meanwhile, the top K accuracy increases with the K tolerance as seen in Table \ref{tab:results:quantitative:topN}.

\subsection{Per-class component detection AP}
Beyond the broad overview of the ranked performances, we present a detailed analysis of per-class performance for each component task.

\begin{table}[]
    \centering
    \caption{Per-class performance on instrument presence detection}
    \label{tab:results:quantitative:instrument}
    \setlength{\tabcolsep}{1.15pt}
    \resizebox{\columnwidth}{!}{%
    \begin{tabular}{@{}lccccccr@{}}
    \toprule
        {Team} &  
        Grasper & \small Bipolar & \small ~Hook~ & \small Scissors & \small Clipper & \small Irrigator & \small Mean\\ 
        \midrule
         SIAT CAMI  & \underline{95.8} & \underline{92.8} & 97.5 & \textbf{94.9} & 81.1 & \textbf{28.7} & \textbf{82.1}\\ 
         Digital Surgery & 94.9 & \textbf{94.2} & \underline{98.4} & \underline{92.8} & \textbf{85.7} & 16.6 & \underline{80.8}\\
         Trequartista & 95.1 & 91.3 & 98.1 & 86.3 & 81.5 & 25.4 & 79.9\\ 
         2AI: Version 2 $^\dag$  & \textbf{96.8} & 88.2 & 98.3 & 88.4 & 81.5 & 23.5 & 79.8\\
         RDV $^\ddag$   & 95.1 & 90.1 & 98.2 & 89.0 & 79.5 & 10.6 & 77.5\\ 
         HFUT-MedIA  & 93.0 & 83.1 & 95.9 & 84.7 & 81.4 & 22.2 & 77.1\\ 
         Attention Tripnet $^\ddag$   & 95.4 & 87.9 & \textbf{98.6} & 88.5 & 78.8 & 10.8 & 77.1\\ 
         Tripnet $^\ddag$   & 86.7 & 82.3 & 97.6 & 79.4 & 80.3 & 19.4 & 74.6\\ 
         Lsgroup  & 91.6 & 85.3 & 96.8 & 76.3 & 76.5 & 14.0 & 73.8\\ 
         ANL Triplet  & 88.3 & 68.6 & 96.6 & 84.0 & \underline{82.4} & 19.8 & 73.6\\
         Casia Robotics  & 92.0 & 88.2 & 97.7 & 67.7 & 72.1 & 15.7 & 72.6\\ 
         SJTU-IMR &    85.7 & 89.7 & 97.4 & 65.7 & 76.6 & 18.7 & 72.6\\ 
         J\&M  &  91.7 & 72.9 & 96.6 & 48.4 & 76.7 & \textbf{28.7} & 69.4\\ 
         Ceaiik & 88.1 & 84.9 & 98.0 & 52.5 & 66.8 & 21.1 & 68.9\\ 
         CITI SJTU   & 92.4 & 92.1 & 66.6 & 62.7 & 78.4 & 12.6 & 67.8\\ 
         Band of Broeders $^\P$  & 91.6 & 55.5 & 94.2 & 66.8 & 66.8 & 03.6 & 63.4\\ 
         SK &    57.3 & 72.5 & 30.6 & 61.5 & 70.6 & 22.3 & 52.6\\ 
         MMLAB & 86.9 & 44.2 & 88.8 & 28.6 & 31.0 & 19.5 & 50.1\\ 
         MTL baseline $^\ddag$   & 81.5 & 58.9 & 93.2 & 13.8 & 37.8 & 04.4 & 48.6\\ 
         2AI: Version 1  &  83.1 & 57.7 & 91.4 & 11.7 & 28.1 & 03.5 & 46.2\\ 
         HFUT-NUS   & 49.8 & 80.0 & 58.3 & 03.3 & 06.5 & 05.9 & 34.1\\ 
         CAMP & 61.2 & 15.6 & 72.0 & 03.5 & 20.4 & 08.9 & 30.4\\ 
         NCT-TSO   & 39.0 & 81.7 & 25.7 & 05.1 & 05.6 & 05.7 & 27.3\\ 
         Med Recognizer & 56.3 & 09.2 & 47.0 & 03.0 & 04.1 & 03.3 & 20.6\\
         \midrule
         \small Mean ± stdev & \scriptsize{82.9±16.7} &\scriptsize{73.6±23.1} & \scriptsize{84.7±22.4} &\scriptsize{56.6±33.2} &\scriptsize{60.4±28.4} &\scriptsize{15.2±8.1} &\scriptsize{62.5±18.9} \\
        \bottomrule 
        \multicolumn{8}{l}{\makecell[l]{\scriptsize{{\bf bold} = best score and \underline{underlined} = second best. Not eligible for award: $\ddag$ organizers' baselines,}\\ \scriptsize{$\dag$ post-challenge submission,  $\P$ used non-public third-party dataset. }}}\\
    \end{tabular}
    }
\end{table}

On surgical instrument presence detection, the most frequently used instruments, {\it hook} and {\it grasper}, are the most correctly detected, as shown in Table \ref{tab:results:quantitative:instrument}. Their recognition APs are above 90\% for half of the methods and their overall mean performances are above 82\%. 
The suction irrigation device ({\it irrigator}) is only used when the field is unclear, resulting in a low usage frequency and mean performance of 15.2\%.
The scissors with the highest standard deviation of ±33.2 is the most complicated instrument to recognize as it often confounded with other instruments such as {\it grasper, bipolar,} and {\it clipper}.

\begin{table*}[!thbp]
    \centering
    \caption{Per-class performance on verb recognition.}
    \label{tab:results:quantitative:verb}
    \setlength{\tabcolsep}{3pt}
    \resizebox{0.9\textwidth}{!}{%
    \begin{tabular}{@{}l*{11}{c}r@{}}
        \hline
        {Team} & \phantom{abc} &
        Grasp & Retract & Dissect & Coagulate & ~~Clip~~ & ~~Cut~~ & Aspirate & Irrigate & ~~Pack~~ & Null & Mean\\ 
        \hline
        Trequartista   &  & 54.0 & \underline{55.5} & \textbf{79.6} & 70.6 & 80.7 & \textbf{81.6} & 20.9 & 01.8 & \underline{48.8} & 30.2 & \textbf{52.9} \\ 
        SIAT CAMI  &  & \underline{56.0} & 46.7 & 69.2 & \textbf{72.8} & 81.2 & \underline{74.2} & \underline{28.1} & 01.9 & 48.7 & \textbf{31.}7 & \underline{51.5} \\ 
        2AI: Version 2 $^\dag$  &  & \underline{56.0} & 46.8 & \underline{78.3} & 68.5 & 80.6 & 72.8 & 17.6 & 01.8 & 44.8 & 28.6 & 50.1 \\ 
        Digital Surgery   &  & 53.2 & 45.7 & 68.3 & 70.6 & \textbf{86.2} & 74.1 & 18.4 & 03.0 & \textbf{49.3} & 26.4 & 50.0 \\ 
        RDV $^\ddag$  &  &  52.4 & 48.5 & 73.2 & 69.5 & 80.2 & 70.5 & 09.9 & 01.2 & 37.1 & 28.2 & 47.5 \\ 
        ANL Triplet  &  & 44.6 & \textbf{60.7} & 73.7 & 62.2 & \underline{82.4} & 69.1 & 11.4 & 00.9 & 37.0 & 26.9 & 47.3 \\ 
        HFUT-MedIA  &  & \textbf{56.8} & 41.8 & 68.2 & 64.0 & 81.6 & 67.2 & 17.2 & \textbf{16.2} & 20.4 & \underline{30.3} & 46.7 \\ 
        J\&M  &  & 47.4 & 44.9 & 75.0 & \underline{71.6} & 77.9 & 36.4 & \textbf{37.8} & \underline{11.2} & 37.7 & 24.3 & 46.7 \\ 
        Lsgroup  &  & 50.3 & 46.0 & 73.5 & 65.1 & 76.6 & 63.1 & 17.3 & 00.5 & 21.1 & 25.3 & 44.3 \\ 
        Casia Robotics  &  & 46.8 & 43.7 & 72.1 & 65.8 & 71.2 & 63.8 & 12.0 & 08.4 & 23.5 & 27.8 & 43.9 \\ 
        Attention Tripnet $^\ddag$  &  & 53.2 & 39.4 & 71.4 & 65.1 & 79.2 & 68.4 & 09.1 & 02.8 & 18.1 & 22.9 & 43.4 \\ 
        Tripnet $^\ddag$  &  & 48.9 & 48.0 & 70.2 & 67.5 & 79.4 & 60.2 & 19.6 & 00.9 & 08.7 & 21.5 & 42.9 \\ 
        SJTU-IMR &   & 57.6 & 47.1 & 74.8 & 69.9 & 76.6 & 43.0 & 14.5 & 00.4 & 10.4 & 26.1 & 42.5 \\ 
        Ceaiik &   & 50.0 & 43.5 & 75.4 & 61.7 & 66.1 & 40.8 & 14.1 & 03.3 & 23.1 & 23.1 & 40.5 \\ 
        CITI SJTU  &  & 54.5 & 45.2 & 72.9 & 66.9 & 67.0 & 04.2 & 12.1 & 00.5 & 20.6 & 23.7 & 37.1 \\ 
        Band of Broeders $^\P$ &  & 44.1 & 35.9 & 68.9 & 55.0 & 67.9 & 52.0 & 02.7 & 00.4 & 01.2 & 20.0 & 35.2 \\ 
        MMLAB  &  & 49.0 & 38.0 & 61.0 & 43.3 & 32.4 & 19.1 & 11.5 & 03.9 & 33.8 & 23.4 & 31.8 \\ 
        SK &   & 47.4 & 23.0 & 53.3 & 50.8 & 70.7 & 02.3 & 17.9 & 00.5 & 12.7 & 22.7 & 30.4 \\ 
        MTL baseline $^\ddag$  &  &  37.7 & 38.7 & 66.6 & 47.2 & 39.7 & 12.5 & 05.0 & 00.3 & 09.3 & 17.9 & 27.8 \\ 
        2AI: Version 1  &  & 43.5 & 34.9 & 58.1 & 39.9 & 28.9 & 13.6 & 02.5 & 00.4 & 01.2 & 18.7 & 24.4 \\ 
        HFUT-NUS  &  & 40.9 & 21.7 & 39.4 & 59.6 & 06.4 & 04.5 & 03.6 & 01.1 & 04.2 & 18.1 & 20.2 \\ 
        CAMP  &  & 42.6 & 26.0 & 45.5 & 21.4 & 23.5 & 06.2 & 06.3 & 00.3 & 04.7 & 16.9 & 19.5 \\ 
        Naive CNN $^\ddag$  &  & 34.4 & 34.2 & 55.5 & 16.4 & 07.8 & 03.4 & 06.3 & 00.8 & 03.4 & 19.0 & 18.3 \\ 
        NCT-TSO  &  & 24.9 & 16.5 & 21.5 & 55.6 & 05.2 & 05.6 & 07.2 & 00.6 & 00.8 & 17.8 & 15.7 \\ 
        Med Recognizer  &  & 35.6 & 21.1 & 32.3 & 07.4 & 01.7 & 01.9 & 02.5 & 00.4 & 01.9 & 21.8 & 12.8 \\
        \midrule
        Mean ± stdev && 47.8±7.8 & 40.0±11.2 & 64.3±15.2 & 58.0±16.3 & 60.1±28.1 & 42.0±29.2 & 13.3±8.5 & 2.6±3.9 & 21.6±16.7 & 23.9±4.3 & 37.7±12.1   \\
        \bottomrule
        \multicolumn{13}{l}{\makecell[l]{\small{
		{\bf bold} = best score and \underline{underlined} = second best. Not eligible for award: $\dag$ post-challenge submission, $\ddag$ organizers' baselines, $\P$ used non-public third-party dataset. }}}\\
    \end{tabular}
    }
\end{table*}

Table \ref{tab:results:quantitative:verb} presents the per-class performance for the verbs.
The most frequently used verbs such as {\it grasp, retract, dissect} are detected above 50.0\% by the top models and above 40\% on average. 
Verbs such as {\it dissect, coagulate, clip, cut}, which have the strongest affinity with a particular instrument class, are detected above 70\% by the top models and with a higher average challenge performance. This confirms that triplets are instrument-centric.
The average performance for {\it cut} is approximately 50\% of the top team score. This is likely affected by the low detection of the performing instrument, {\it scissors}.
In cases where an instrument has multiple frequent verbs, the performance tends to spread out over those verbs according to their prevalence: for example {\it retract $\approx$ grasp $\gg$ pack} for {\it grasper}, {\it aspirate $\gg$ irrigate} for {\it irrigator}, etc. 
{\it Irrigate} is the least detected verb, likely due to its temporal nature as it can only be distinguished from {\it aspirate} based on the temporal dynamics of the fluid. Remarkably, team HFUT-MedIA leveraged graph convolution networks, notable for temporal action detection, to better detect this verb.

\begin{table*}[!thb]
    \centering
    \caption{Per-class performance on target recognition.{$^\S$}}
    \label{tab:results:quantitative:target}
    \setlength{\tabcolsep}{5pt}
    \resizebox{\textwidth}{!}{%
    \begin{tabular}{l@{}l*{12}{c}r@{}}
        \toprule
        {Team} & \phantom{abc} &
        ~\rotatebox[origin=l]{85}{Gallbladder}~~ & 
        ~~\rotatebox[origin=l]{85}{Cystic-duct}~~ & 
        ~~\rotatebox[origin=l]{85}{Cystic-artery}~~ & 
        ~~\rotatebox[origin=l]{85}{Blood-vessel}~~ & 
        ~~\rotatebox[origin=l]{85}{Fluid} & 
        \rotatebox[origin=l]{85}{\makecell[l]{Abdominal-wall\\or cavity}} & 
        \rotatebox[origin=l]{85}{Liver} & 
        ~~~\rotatebox[origin=l]{85}{Omentum}~~~ & 
        ~~\rotatebox[origin=l]{85}{Peritoneum}~~ & 
        ~~\rotatebox[origin=l]{85}{Gut}~~ & 
        ~~~\rotatebox[origin=l]{85}{Specimen-bag}~~~ & 
        \rotatebox[origin=l]{85}{Null} & 
        \rotatebox[origin=l]{85}{Mean} \\  
        \midrule
        Trequartista   &  & \underline{91.4} & \underline{66.8} & 30.6 & \textbf{33.0} & 20.9 & \underline{60.8} & \textbf{82.5} & 00.7 & \underline{37.5} & 08.4 & \underline{89.2} & 30.2 & \textbf{46.4} \\ 
        SIAT CAMI  &  & 89.4 & 64.3 & \underline{31.2} & 15.4 & \underline{28.1} & \textbf{67.3} & 79.5 & 00.6 & 32.6 & \underline{16.3} & 85.4 & \textbf{31.7} & \underline{45.5} \\ 
        2AI: Version 2 {$^\dag$} &  & 89.4 & \textbf{67.3} & 26.1 & \textbf{33.0} & 17.6 & 43.1 & 75.0 & 00.8 & 24.4 & 14.8 & \underline{89.2} & 28.6 & 42.8 \\ 
        Digital Surgery   &  & \textbf{93.3} & 61.2 & 29.4 & 02.1 & 18.4 & 30.9 & \textbf{82.5} & 00.6 & \textbf{39.9} & 10.5 & \textbf{92.6} & 26.4 & 41.1 \\ 
        ANL TRIPLET   &  & 86.5 & 65.5 & 26.9 & 27.4 & 11.4 & 35.9 & 69.9 & 00.4 & 33.9 & \textbf{18.3} & 78.6 & 26.9 & 40.5 \\ 
        J\&M   &  & 87.3 & 43.1 & 27.3 & 03.8 & \textbf{37.8} & 56.7 & 83.0 & 00.3 & 14.5 & 15.2 & 73.6 & 24.3 & 39.2 \\ 
        HFUT-MedIA   &  & 85.0 & 61.0 & 24.5 & 25.0 & 17.2 & 26.5 & 69.1 & 00.9 & 15.5 & 08.1 & 86.1 & \underline{30.3} & 37.8 \\ 
        RDV {$^\ddag$} &  & 88.2 & 54.7 & \textbf{32.0} & 18.3 & 09.9 & 27.7 & 70.9 & 00.5 & 25.2 & 08.4 & 83.5 & 28.2 & 37.7 \\ 
        Lsgroup   &  & 88.3 & 51.4 & 21.6 & 22.6 & 17.3 & 24.9 & 63.4 & 00.5 & 20.4 & 08.1 & 70.7 & 25.3 & 34.9 \\ 
        CITI SJTU   &  & 84.8 & 55.2 & 23.6 & 04.8 & 12.1 & 44.8 & 64.7 & 01.0 & 06.7 & 13.5 & 78.1 & 23.7 & 34.8 \\ 
        SJTU-IMR   &  & 83.6 & 54.7 & 24.7 & 08.9 & 14.5 & 34.2 & 67.0 & 00.8 & 09.6 & 05.2 & 76.2 & 26.1 & 34.1 \\ 
        Tripnet {$^\ddag$} &  & 82.2 & 49.4 & 24.7 & 07.0 & 19.6 & 16.0 & 68.6 & 01.2 & 09.1 & 02.4 & 80.6 & 21.5 & 32.2 \\ 
        MMLAB  &  & 79.6 & 45.3 & 18.1 & 14.2 & 11.5 & 10.4 & 51.3 & 00.6 & 26.9 & 16.0 & 77.6 & 23.4 & 31.6 \\ 
        Casia Robotics  &  & 84.3 & 48.9 & 16.7 & 05.9 & 12.0 & 21.2 & 59.5 & \textbf{03.9} & 06.0 & 12.1 & 71.8 & 27.8 & 31.2 \\ 
        Ceaiik & & 84.3 & 44.7 & 21.0 & 05.4 & 14.1 & 21.4 & 55.1 & 01.2 & 12.0 & 08.7 & 76.0 & 23.1 & 30.9 \\ 
        Attention Tripnet {$^\ddag$} &  & 77.4 & 42.5 & 23.1 & 13.8 & 09.1 & 22.2 & 41.0 & \underline{02.0} & 18.5 & 03.6 & 79.7 & 22.9 & 30.0 \\ 
        Band of Broeders $^\P$ &  & 83.9 & 42.6 & 08.2 & 01.7 & 02.7 & 03.7 & 74.6 & 00.3 & 03.7 & 01.1 & 67.1 & 20.0 & 26.2 \\ 
        2AI: Version 1  &  & 76.9 & 22.1 & 05.8 & 01.7 & 02.5 & 00.9 & 45.6 & 00.3 & 03.7 & 01.1 & 63.7 & 18.7 & 20.5 \\ 
        SK &  & 71.5 & 33.4 & 12.4 & 03.9 & 17.9 & 03.5 & 24.9 & 01.1 & 02.7 & 03.5 & 42.4 & 22.7 & 20.2 \\ 
        MTL baseline $^\ddag$  &  & 76.2 & 21.3 & 11.2 & 01.9 & 05.0 & 03.0 & 42.5 & 00.3 & 04.9 & 02.2 & 48.3 & 17.9 & 19.8 \\ 
        HFUT-NUS  &  & 52.6 & 12.3 & 05.2 & 03.1 & 03.6 & 03.8 & 40.9 & 01.1 & 04.7 & 02.5 & 11.7 & 18.1 & 13.5 \\ 
        NCT-TSO  &  & 40.9 & 10.6 & 12.1 & 16.2 & 07.2 & 00.9 & 33.7 & 00.3 & 02.5 & 00.7 & 02.8 & 17.8 & 12.3 \\ 
        CAMP  &  & 58.6 & 10.8 & 05.0 & 02.7 & 06.3 & 01.5 & 14.6 & 00.6 & 02.6 & 04.3 & 16.6 & 16.9 & 11.8 \\ 
        Med Recognizer  &  & 56.4 & 12.0 & 04.9 & 01.6 & 02.5 & 00.9 & 10.1 & 00.5 & 03.0 & 00.7 & 05.3 & 21.8 & 10.1 \\
        \midrule
        Mean ± stdev && 78.8±13.5 & 43.4±19.0 & 19.4±9.1 & 11.4±10.3 & 13.3±8.5 & 23.4±20.3 & 57.1±21.3 &	0.9±0.7 & 15.0±12.3 & 7.7±5.7 & 64.5±27.9 &	23.9±4.3 & 30.2±11.0 \\
        \bottomrule 
        \multicolumn{15}{l}{\makecell[l]{\small{
		{\bf bold} = best score and \underline{underlined} = second best. Not eligible for award: $\dag$ post-challenge submission, $\ddag$ organizers' baselines, $\P$ used non-public third-party dataset. $\S$ shows only the targets in test videos.}}}\\
    \end{tabular}
    }
\end{table*}

Finally, we analyze per-class performance on target recognition which appears to be the most challenging component to correctly detect. As shown in Table \ref{tab:results:quantitative:target}, the {\it gallbladder} and {\it specimen-bag} are the most recognized targets, with the top models exceeding  $90.0\%$ AP. Their average performance across all methods is above 64.0\%. 

\begin{table*}[!thb]
    \centering
    \caption{Wilcoxon signed-rank test of the competing teams for rank stability.}
    \label{tab:results:stability}
    \resizebox{0.9\textwidth}{!}{%
        \includegraphics[width=1.0\linewidth]{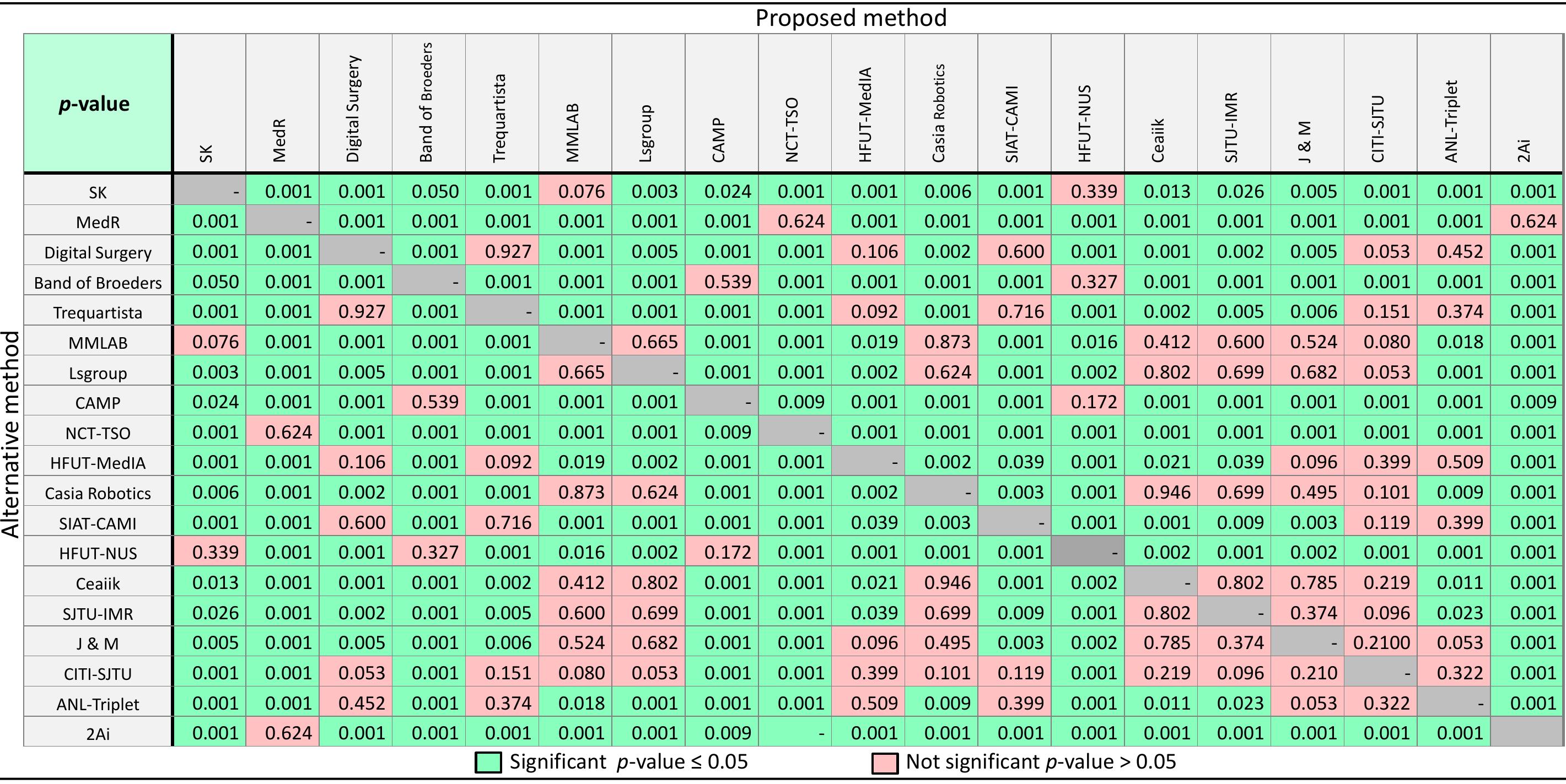}
    }
\end{table*}

Other targets such as {\it liver, cystic-duct, abdominal wall} and {\it cavity}, are moderately detected. This is likely due to their obvious nature and clearer boundaries compared to the less detected ones. Interactions with them are easier to ascertain than interactions with much smaller structures such as {\it cystic-artery} and other {\it blood-vessels}. Within the {\it cystic-pedicle}, the {\it cystic-duct} is the most detected tubular structure. The {\it cystic artery} in itself is hard to differentiate from other {\it blood-vessels}. This shows how deceptively complicated the task of anatomical target detection could be.
The {\it peritoneum}, which covers the entire cavity, appears as a transparent layer making it difficult to identify. The heavily super-classed {\it omentum}, {\it peritoneum}, and {\it gut} are the least detected in this case.

{\red For single-task objectives, models proposed by SIAT-CAMI and Digital Surgery have the best performances and seem the most suitable for surgical instrument recognition. Both teams utilized a very deep feature extraction backbone and multi-scale aggregation of local and global features for final classification. 
On the other hand, models presented by Trequartista, SIAT-CAMI, and 2AI would be the best bet for modeling surgical action recognition which could be attributed to their advanced exploitation of the video temporal information. They also show promising performances on the target recognition task. }

{\red Taking all the per-component analysis into account, it is observed that the target recognition rate has a larger impact on the overall triplet recognition. This is most likely due to the target being the most challenging component to be correctly recognized and classified by the presented models. A model would have a higher probability of recognizing the other components if it can correctly recognize the complex targets.

For the joint task, there is no clear trade-off in the sub-task modeling, instead, balancing their scores guarantees a better overall triplet recognition performance as is the case in the 2AI version 2 model (Tables \ref{tab:results:quantitative:instrument},\ref{tab:results:quantitative:verb},\ref{tab:results:quantitative:target} and \ref{tab:results:quantitative:summary}).
}

\subsection{Result ranking stability }
To measure the rank stability, we employ Wilcoxon signed-rank test as a non-parametric alternative to the dependent samples t-test since our data is not multivariate normal and teams' predictions could present many outliers.
Using this, we test each team's method against a null hypothesis ($H_0$) to ascertain the statistical significance of its performance over others.
The $H_0$ states that the difference between the proposed method and the alternative methods has a mean signed-rank of zero. Each $H_0$ test of one team against another produces a $p$-value between 0 and 1 as a measure of its level of statistical significance with a smaller value showing stronger evidence to reject the null hypothesis.
Typically, a $p\leq 0.05$ is considered statistically significant.

To perform this analysis, we sample $N=30$ random batches of $100$ consecutive frames to simulate a video clip evaluation and perform the Wilcoxon test by iteratively using each competing team as a proposed method against the rest of the teams as its alternatives. The obtained $p$-values, tabulated in Table \ref{tab:results:stability}, show that closely-ranked teams do not significantly improve each other methods (accept $H_0$), whereas distantly ranked teams show a significant difference in their model performances (reject $H_0$).
At a 5\% confidence level, we conclude that the CholecTriplet2021 challenge has assembled teams with both similar and diverse methods in both modeling and performance.

\subsection{Model ensemble results }

\begin{table*}[!thb]
    \centering
    \caption{Performance summary of the ensemble methods in comparison with the top 7 methods at the challenge.}
    \label{tab:results:quantitative:ensemble_summary}
    \setlength{\tabcolsep}{10pt}
    \resizebox{.9\textwidth}{!}{%
    \begin{tabular}{@{}llclcrrlcr@{}}
    \toprule
        & \multirow{2}{*}{Method} &  \phantom{abc} &
        \multicolumn{3}{c}{Component detection} & \phantom{abc} &
        \multicolumn{3}{c}{Triplet association}\\ 
        \cmidrule{4-6} 
        \cmidrule{8-10} 
        &&& $AP_{I}$ & $AP_{V}$ & $AP_{T}$ && $AP_{IV}$ & $AP_{IT}$ & $AP_{IVT}$ \\ 
        \midrule
        \multirow{7}{*}{Top Challenge Teams} 
        & Trequartista && 79.9 & \underline{52.9} & \bf 46.4 && 39.0 & 41.9 & 38.1\\ 
        & SIAT CAMI && {82.1} & 51.5 & \underline{45.5} && 37.1 & 43.1 & 35.8\\ 
        & HFUT-MedIA && 77.1 & 46.7 & 37.8 && 33.1 & 35.9 & 32.9\\ 
        & RDV $^\ddag$ && 77.5 & 47.5 & 37.7 && 39.4 & 39.6 & 32.7\\ 
        & CITI SJTU && 67.8 & 37.1 & 34.8 && 29.9 & 33.0 & 32.0\\ 
        & ANL Triplet && 73.6 & 47.3 & 40.5 && 32.6 & 37.1 & 31.9\\ 
        & Digital Surgery && 80.8 & 50.0 & 41.1 && 35.1 & 35.7 & 31.7\\ 
        \midrule
        \multirow{6}{*}{Model Ensemble } 
        & Averaging  && \underline{82.4} & \underline{52.9} & 44.7 && \underline{40.4}  & \underline{43.5}  & 38.9 \\ 
        & Weighted Averaging   && \bf 82.5 & \bf 53.1 & 44.9  && \bf 40.5  & \bf 43.8  & 39.2 \\  
        & Soft Voting   && 79.6  & 46.7 & 42.6  && 35.3 & 39.8 & 35.1 \\
        & Deep ensemble && 71.4  & 37.3  & 28.6  && 30.5  & 30.5  &  30.3 \\  
        & Deep weighted ensemble   && 81.9  &  51.5 & 44.0  && 39.3  & 43.1  & \underline{40.5} \\  
        & Deep per-class weighted ensemble  && 81.4  &  52.2 & \bf 46.4  && 40.0  &  42.9 & \bf 42.4 \\  
        \bottomrule
        \multicolumn{9}{l}{\makecell[l]{\small{
		{\bf bold} = best score and \underline{underlined} = second best. $\ddag$ = organizers' baseline}}}\\
    \end{tabular}
    }
\end{table*}

Table \ref{tab:results:quantitative:ensemble_summary} demonstrates that combining decisions from multiple models indeed helps to improve their overall performance. 
A simple averaging leads to an additional 0.8\% gain in triplet recognition AP and +1.1\% AP when the models' contributions are weighted by their individual strengths.
Although an ensemble builds a strong learner from a group of weak learners, voting is surprisingly ineffective in this regard.
In comparison to non-trainable alternatives, training ensemble techniques help to better minimize noise, bias, and variance errors, resulting in superior performances. 
Learning a new averaging weight appears more efficient than training totally new prediction models (Deep ensemble), as shown in Table \ref{tab:results:quantitative:ensemble_summary}.
Learning the model weights per task category rather than general class weights per model is even more fascinating and reaches the highest mAP of 42.4\% for surgical action triplet recognition. 
In general, performance improvement with the model ensemble is obtained across all the six sub-tasks.

\begin{figure*}[!thbp]
    \includegraphics[width=1.0\linewidth]{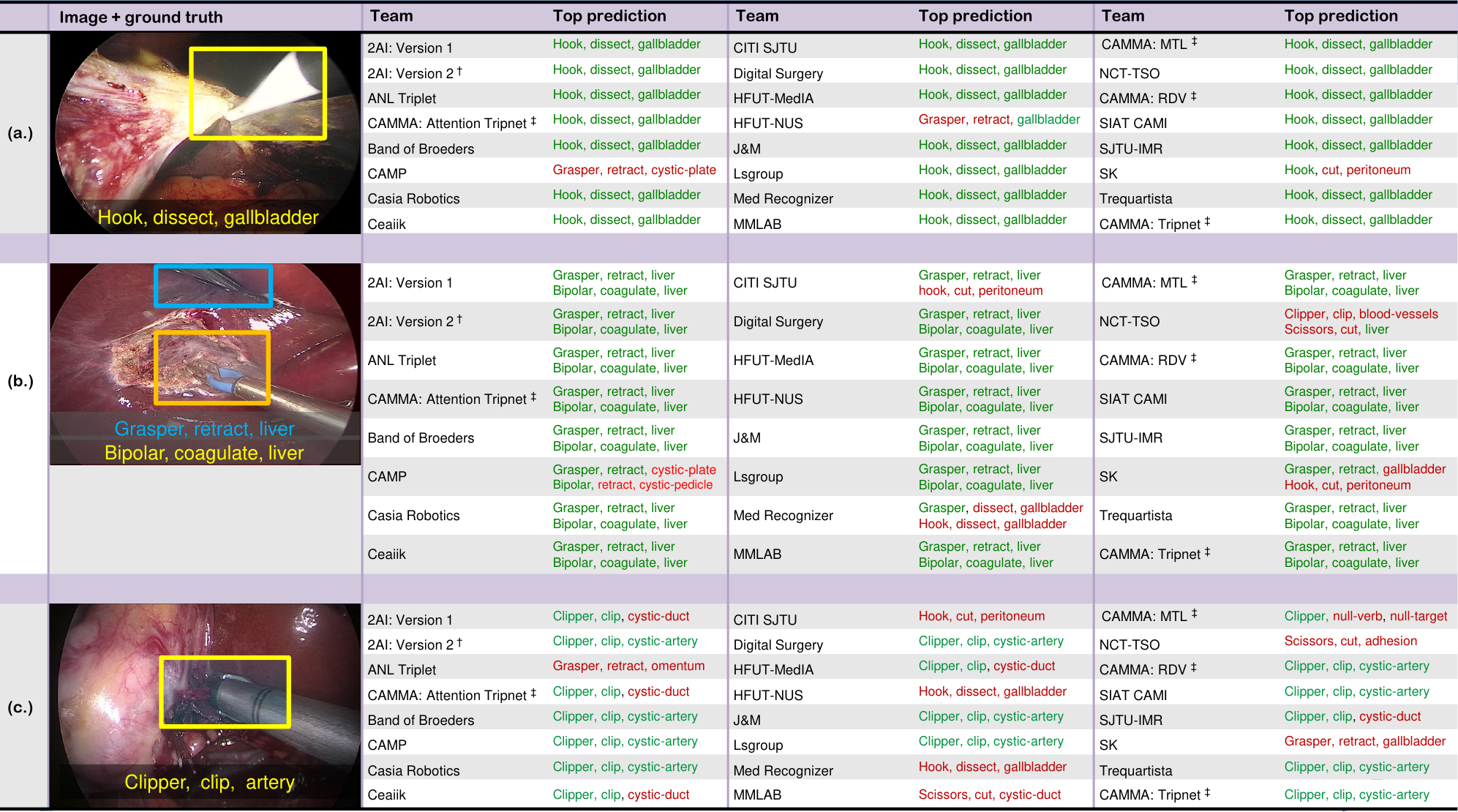} 
    \caption{Qualitative results visualizing triplet predictions:  {\normalfont a cross-section of teams' top $k$ predictions on an input image depicting $k$ action triplets: (a.) easy case, (b.) moderate case, and (c.) difficult case. $\dag$ = post-challenge submission, $\ddag$ = organizers' baselines, $\P$ = used non-public third-party dataset. {\green green} = incorrect prediction {\red red} = incorrect prediction.}}
    \label{tab:results:qualitative:topk}
\end{figure*}

\subsection{Qualitative results}
To better analyze the quality of the detections, we visualize the top K predictions for each model on sample frames with K triplet instances.
An easy example in Fig. \ref{tab:results:qualitative:topk} showcases an image frame with a triplet of the best-recognized instrument ({\it hook}), the best recognized verb ({\it dissect}), and best-recognized target ({\it gallbladder}). This case is correctly detected by $\approx88\%$ of the teams.  A moderately difficult case of a frame with multiple triplets recorded seven incorrect predictions.
An image frame showing a single triplet involving a {\it cystic-artery} proves to be difficult for more than half of the teams. The incorrect prediction is mostly on the target component of the triplet.

\begin{table*}[!thbp]
    \centering
    \caption{Qualitative results regarding the quality of triplet recognition on the CholecT50 dataset. \normalfont The results are obtained on a concatenation of 3 different surgical videos from the testing set. The methodologies employed by each model are indicated in columns 1-7. {\red The triplet flows are categorized by their instrument, and the color shades illustrate their varying interactions (verbs) on different targets.}}
    \label{tab:results:qualitative:workflow}
    \resizebox{1.0\textwidth}{!}{%
    \begin{tabular}{
    @{}p{0.15\textwidth}@{}
    p{0.005\textwidth}p{0.005\textwidth}p{0.005\textwidth}p{0.005\textwidth}p{0.005\textwidth}p{0.005\textwidth}p{0.005\textwidth}P{0.001\textwidth}
    P{0.11\textwidth}@{}P{0.1\textwidth}p{0.08\textwidth}P{0.09\textwidth}P{0.09\textwidth}P{0.09\textwidth}@{}
    } \toprule
    \multirow{2}{*}{\makecell{\\TEAM}} & 
    \parbox[t]{0mm}{\multirow{2}{*}{\rotatebox[origin=r]{90}{Multi-task Learning}}} & 
    \parbox[t]{0mm}{\multirow{2}{*}{\rotatebox[origin=r]{90}{Temporal Modeling}}} & 
    \parbox[t]{0mm}{\multirow{2}{*}{\rotatebox[origin=r]{90}{Attention Mechanism}}} & 
    \parbox[t]{0mm}{\multirow{2}{*}{\rotatebox[origin=r]{90}{Graph Convolution}}} & 
    \parbox[t]{0mm}{\multirow{2}{*}{\rotatebox[origin=r]{90}{Ensemble Methods}}} & 
    \parbox[t]{0mm}{\multirow{2}{*}{\rotatebox[origin=r]{90}{+ Phase Labels}}} & 
    \parbox[t]{0mm}{\multirow{2}{*}{\rotatebox[origin=r]{90}{+ Spatial Labels}}} & &
    \multicolumn{6}{c}{\makecell[c]{\\Action Triplets Performed Using Instrument:}} \\\\
    \cmidrule{10-15} \cmidrule{1-1}\\
    &&&&&&&&& Grasper & {Bipolar} & {Hook} & {Scissors} & {Clipper} & {Irrigator} \\
    \\[0.1in]
    Groundtruth &  & & & & & & & &
    \multicolumn{6}{r}{
    \begin{minipage}{0.66\textwidth}
      \includegraphics[width=1.0162\linewidth, height=6mm]{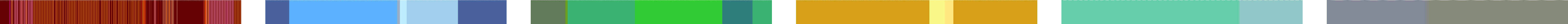}%
    \end{minipage}%
    }\\\midrule 
    Trequartista & \OK &&& & \OK & \OK & & &
    \multicolumn{6}{l}{
    \begin{minipage}{0.66\textwidth}
      \includegraphics[width=1.0162\linewidth, height=6mm]{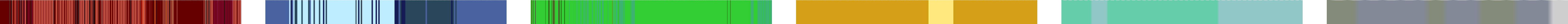}
    \end{minipage}
    }\\\midrule 
    
    2AI (version 2)$^\dag$ & \OK & \OK & & & \OK & \OK & & &
    \multicolumn{6}{l}{
    \begin{minipage}{0.66\textwidth}
      \includegraphics[width=1.0162\linewidth, height=6mm]{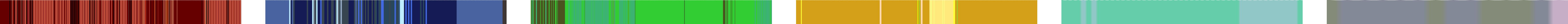}
    \end{minipage}
    }\\\midrule 
    
    SIAT-CAMI & \OK &\OK & \OK &&&&&&
    \multicolumn{6}{l}{
    \begin{minipage}{0.66\textwidth}
      \includegraphics[width=1.0162\linewidth, height=6mm]{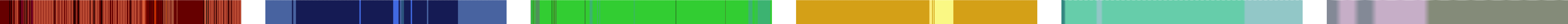}
    \end{minipage}
    }\\\midrule 
    
    HFUT-MedIA & \OK &&&\OK&&&&&
    \multicolumn{6}{l}{
    \begin{minipage}{0.66\textwidth}
      \includegraphics[width=1.0162\linewidth, height=6mm]{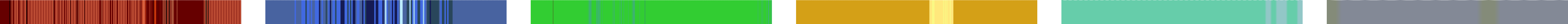}
    \end{minipage}
    }\\\midrule 
    
    RDV$^\ddag$ & \OK & & \OK&& &&&&
    \multicolumn{6}{l}{
    \begin{minipage}{0.66\textwidth}
      \includegraphics[width=1.0162\linewidth, height=6mm]{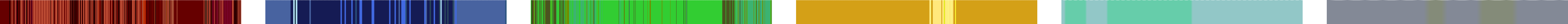}
    \end{minipage}
    }\\\midrule 
    
    CITI SJTU & \OK &\OK & && &&&&
    \multicolumn{6}{l}{
    \begin{minipage}{0.66\textwidth}
      \includegraphics[width=1.0162\linewidth, height=6mm]{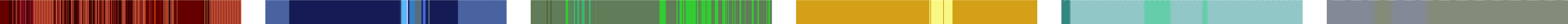}
    \end{minipage}
    }\\\midrule 
    
    ANL Triplet & \OK &\OK & && &&&&
    \multicolumn{6}{l}{
    \begin{minipage}{0.66\textwidth}
      \includegraphics[width=1.0162\linewidth, height=6mm]{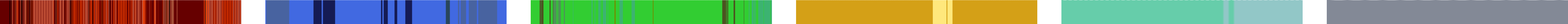}
    \end{minipage}
    }\\\midrule 
    
    Digital Surgery & &\OK & && \OK &&&&
    \multicolumn{6}{l}{
    \begin{minipage}{0.66\textwidth}
      \includegraphics[width=1.0162\linewidth, height=6mm]{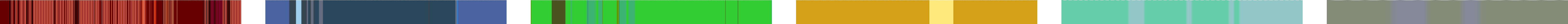}
    \end{minipage}
    }\\\midrule 
    
    Casia Robotics &\OK & & \OK&&  &&&&
    \multicolumn{6}{l}{
    \begin{minipage}{0.66\textwidth}
      \includegraphics[width=1.0162\linewidth, height=6mm]{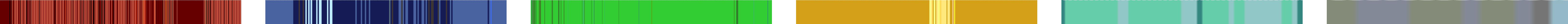}
    \end{minipage}
    }\\\midrule 
    
    Lsgroup & \OK &\OK & &&&&&&
    \multicolumn{6}{l}{
    \begin{minipage}{0.66\textwidth}
      \includegraphics[width=1.0162\linewidth, height=6mm]{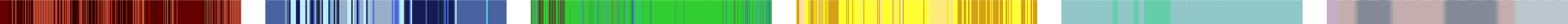}
    \end{minipage}
    }\\\midrule 
    
    J\&M &&\OK & &&&&&&
    \multicolumn{6}{l}{
    \begin{minipage}{0.66\textwidth}
      \includegraphics[width=1.0162\linewidth, height=6mm]{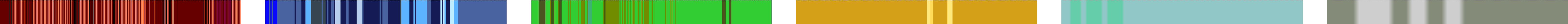}
    \end{minipage}
    }\\\midrule 
    
    Attention Tripnet$^\ddag$ & \OK && \OK &&&&&&
    \multicolumn{6}{l}{
    \begin{minipage}{0.66\textwidth}
      \includegraphics[width=1.0162\linewidth, height=6mm]{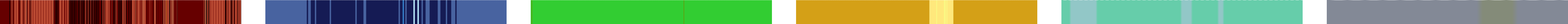}
    \end{minipage}
    }\\\midrule 
    
    Ceaiik & \OK &\OK & &&&&&&
    \multicolumn{6}{l}{
    \begin{minipage}{0.66\textwidth}
      \includegraphics[width=1.0162\linewidth, height=6mm]{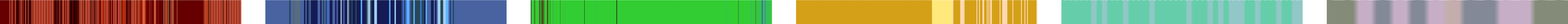}
    \end{minipage}
    }\\\midrule 
    
    SJTU-IMR & \OK &\OK & \OK &&&&&&
    \multicolumn{6}{l}{
    \begin{minipage}{0.66\textwidth}
      \includegraphics[width=1.0162\linewidth, height=6mm]{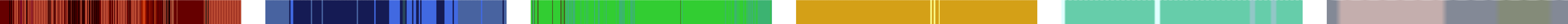}
    \end{minipage}
    }\\\midrule 
    
    Tripnet$^\ddag$ & \OK &&&&&&&&
    \multicolumn{6}{l}{
    \begin{minipage}{0.66\textwidth}
      \includegraphics[width=1.0162\linewidth, height=6mm]{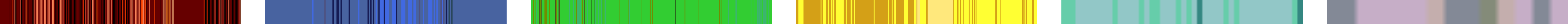}
    \end{minipage}
    }\\\midrule 
    
    SK & \OK &&\OK  &&&&&&
    \multicolumn{6}{l}{
    \begin{minipage}{0.66\textwidth}
      \includegraphics[width=1.0162\linewidth, height=6mm]{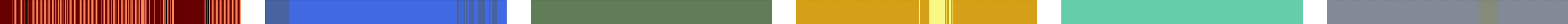}
    \end{minipage}
    }\\\midrule 
    
    MMLAB &&& &\OK &&&&&
    \multicolumn{6}{l}{
    \begin{minipage}{0.66\textwidth}
      \includegraphics[width=1.0162\linewidth, height=6mm]{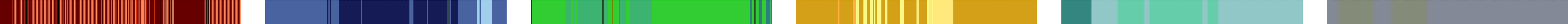}
    \end{minipage}
    }\\\midrule 
    
    Band of Broeders$^\P$ & \OK && &&&& \OK &&
    \multicolumn{6}{l}{
    \begin{minipage}{0.66\textwidth}
      \includegraphics[width=1.0162\linewidth, height=6mm]{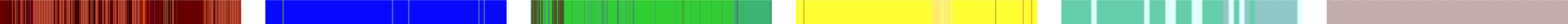}
    \end{minipage}
    }\\\midrule 
    
    MTL Baseline$^\ddag$ & \OK && &&&&&&
    \multicolumn{6}{l}{
    \begin{minipage}{0.66\textwidth}
      \includegraphics[width=1.0162\linewidth, height=6mm]{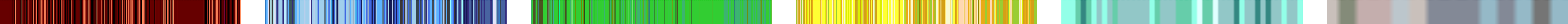}
    \end{minipage}
    }\\\midrule 
    
    NCT-TSO & \OK & & && &&&&
    \multicolumn{6}{l}{
    \begin{minipage}{0.66\textwidth}
      \includegraphics[width=1.0162\linewidth, height=6mm]{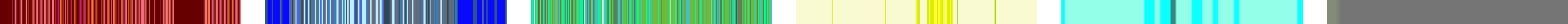}
    \end{minipage}
    }\\\midrule 
    
    2Ai & \OK &\OK & && \OK &\OK&&&
    \multicolumn{6}{l}{
    \begin{minipage}{0.66\textwidth}
      \includegraphics[width=1.0162\linewidth, height=6mm]{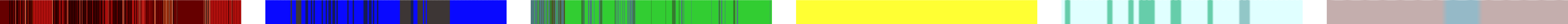}
    \end{minipage}
    }\\\midrule 
    
    HFUT-NUS & \OK & & && &&&&
    \multicolumn{6}{l}{
    \begin{minipage}{0.66\textwidth}
      \includegraphics[width=1.0162\linewidth, height=6mm]{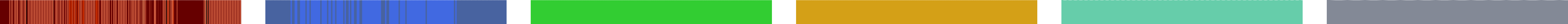}
    \end{minipage}
    }\\\midrule 
    
    CAMP & \OK &\OK & &&& \OK &&&
    \multicolumn{6}{l}{
    \begin{minipage}{0.66\textwidth}
      \includegraphics[width=1.0162\linewidth, height=6mm]{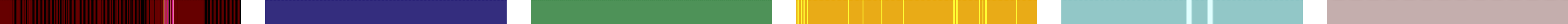}
    \end{minipage}
    }\\\midrule 
    
    Med Recognizer & \OK &\OK & && \OK &&&&
    \multicolumn{6}{l}{
    \begin{minipage}{0.66\textwidth}
      \includegraphics[width=1.0162\linewidth, height=6mm]{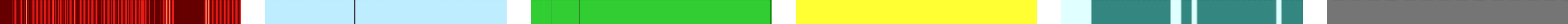}
    \end{minipage}
    }\\\midrule 
    \multicolumn{15}{l}{\makecell[l]{\small{
		Not eligible for award: $\dag$ post-challenge submission, $\ddag$ organizers' baselines, $\P$ used non-public third-party dataset. }}}\\
    \bottomrule
    \end{tabular}
    }
\end{table*}

Finally, we analyze the triplet recognition showing a sequential flow of tool-activity recognition. Here, we group all the triplets by their instruments and use a range of colors to show their variations and transitions in temporal order. The ground-truth flow shows the level of simplicity and complexity of many triplets with regard to their affinities with instruments.
As shown in Table \ref{tab:results:qualitative:workflow}, some triplets can be easily inferred by the instrument information such as scissors (likely {\it cutting} either {\it cystic-artery, cystic-duct, blood-vessels} or some adhesion), clipper (likely clipping the respective anatomical targets). 
However, instruments such as {\it grasper, hook, bipolar,} etc, perform multiple triplets which are not easily deduced from instrument presence alone. The overcrowded coloring of \textit{grasper} shows that it is used for various actions/targets and often for short sequences/intervals.
To better understand the model's capacity to consistently approximate the ground-truth labels, we present their utilized methodologies in the first 8 columns. 
Even with the complexity of actions performed by the grasper, it is better recognized by top methods.
Triplets with bipolar appear very complex to all the methods. A closer attempt is by team ANL-Triplet which leveraged temporal modeling.
Triplets with the hook are less difficult, however, their transition is mostly unclear to all the models. 
Triplets with scissors are the best-approximated predictions due to their limited variability. Specifically, models utilizing attention and/or temporal modeling are the best at this triplet. The use of phase labels mimics methods utilizing temporal modeling.
Triplets with the clipper are another class with limited variability and are closely approximated by most of the models.
Triplets with the irrigator, although limited in variability, appear very confusing to all the presented methods.

\subsection{Limitations }
{%
This challenge and analysis present several limitations, which should be addressed in future iterations. Firstly, as with most challenges, given that the participants are not constrained in terms of modeling, comparisons made between different modeling approaches must be treated as indications rather than facts. Participating submissions employed varying degrees of focus on components such as hyperparameter tuning, thus limiting their comparability. We also note that different submissions use a wide range of parameter counts, potentially due to resource constraints or research priorities, which could greatly affect model performance. Another limitation of this challenge was the enforcement of the causality constraint. Participants were asked to ensure that their submission only made use of past frames to make predictions; however, enforcing this at inference time is computationally impractical as each input image must be processed individually along with its entire temporal context to ensure that there is no leakage of information. Alternative strategies could have been to decide on a fixed and limited temporal context (n frames) that would be used to make a prediction or to allow acausal predictions. Bearing in mind that the value of these systems may lie in real-time systems, we opted to allow an unlimited past context and designed a hidden subset of the test set that was used to test causality.
}

%% file: main/08-conclusion.tex
\section{Conclusion }
As the finest-grained and most comprehensive description of surgical activities for computer vision, surgical action triplets carry significant clinical value. Before this challenge, however, this problem received little attention from the community; in that regard, \textit{CholecTriplet2021} was a success. With a record-high 19 submissions, three of which surpassed the state of the art, this event featured a diverse range of approaches, providing a solid methodological foundation for future research efforts. Most importantly, this challenge showed that surgical action triplet recognition remains an open challenge, with several promising directions to explore. The use of attention, already studied before \textit{CholecTriplet2021}, can be expanded, as shown by several submissions. This challenge also saw the very first uses of graph convolutions and temporal models for surgical action triplets.

Finally, future work should focus on refining the spatial aspect of this problem: locating action triplets in addition to simply detecting their presence would provide much richer information on the surgery. In that sense, full bounding box annotations would be a considerable step forward for research on tool-tissue interaction.
{\red %
Owing to the similarity of tissue manipulation across surgical procedures, action triplet recognition can be adapted to other surgical procedures by following the triplet labeling formalism to annotate the specific procedure data. Transfer learning from existing triplet models or pretraining on the CholecT50 dataset could potentially benefit model convergence on the new data.
}

%% file: main/09-acknowledgement.tex
\section*{Acknowledgment }
{\small
The organizers would like to thank the IHU and IRCAD research teams for their help with the initial data annotation during the CONDOR project. We also thank the EndoVis 2021 organizing committee for providing the platform for this challenge. Specifically, we thank Stefanie Speidel, Lena Maier-Hein, and Danail Stoyanov.}

\section*{Funding }
{\small
This work was supported by French state funds managed within the Investissements d\textsc{\char13}Avenir program by BPI France (project CONDOR) and by the ANR under references: Labex CAMI [ANR-11-LABX-0004], DeepSurg [ANR-16-CE33-0009], IHU Strasbourg [ANR-10-IAHU-02], and National AI Chair AI4ORSafety [ANR-20-CHIA-0029-01].
Software validation and evaluation were performed with servers managed by CAMMA, as well as HPC resources from Unistra Mésocentre and GENCI-IDRIS [Grant 2021-AD011011638R1, 2021-AD011011640R1].
Awards for the challenge were sponsored by NVIDIA and Medtronic plc.

Participating teams would like to acknowledge the following funding: 
{\it\bf NCT-TSO}: Federal Ministry of Health, Germany (BMG) - part of the SurgOmics project.
{\it\bf CAMP}: Ph.D. Fellowship at CAMP Chair at TU Munich.
{\it\bf CASIA Robotics}: National Key Research and Development Program of China [Grant 2017YFB1302704], the National Natural Science Foundation of China [Grant U1713220], the Beijing Science and Technology Plan [Grant Z191100002019013], and the Youth Innovation Promotion Association of the Chinese Academy of Sciences [Grant 2018165].
{\it\bf HFUT-MedIA}: National Natural Science Foundation of China [No. 91846107].
{\it\bf SIAT-CAMI}: Guangdong Key Area Research and Development Program [2020B010165004] and Shenzhen Key Basic Science Program [JCYJ20180507182437217].
{\it\bf MMLAB}: Shun Hing Institute of Advanced Engineering [SHIAE project \#BME-p1-21, 8115064] at the Chinese University of Hong Kong (CUHK) and Singapore Academic Research Fund [Grant R397000353114]. 
{\it\bf CITI-SJTU and SJTU-IMR:} Shanghai Municipal Science and Technology  Commission [20511105205] and National Natural Science of China [U20A20199].
{\it\bf 2AI}: NORTE-01-0145-FEDER-000045” and "NORTE-01-0145-FEDER-000059", supported by Northern Portugal Regional Operational Programme (NORTE 2020), under the Portugal 2020 Partnership Agreement, through the European Regional Development Fund (FEDER), FCT and FCT/MCTES [project: UIDB/05549/2020, UIDP/05549/2020]. Also, Fundação para a Ciência e a Tecnologia (FCT), Portugal and the European Social Found, European Union, for funding support through the “Programa Operacional Capital Humano” POCH [PhD Grants SFRH/BD/136721/2018, SFRH/BD/136670/2018].
{\it\bf HFUT-NUS}: the Chinese Scholarship Council [No. 202006690025].
{\it\bf Lsgroup}: Ministry of Science and Technology of the People´s Republic of China [2021ZD0201900, 2021ZD0201903].
{\it\bf Digital Surgery}:
Wellcome/EPSRC Centre for Interventional and Surgical Sciences (WEISS) [203145Z/16/Z]; Engineering and Physical Sciences Research Council (EPSRC) [EP/P027938/1, EP/R004080/1, EP/P012841/1]; The Royal Academy of Engineering Chair in Emerging Technologies scheme.
}

%% file: credit.tex
\newpage
\onecolumn
\section*{CRediT authorship contribution statement}
\vspace{10pt}

\noindent{\bf C.I. Nwoye} : Conceptualization, Data Curation, Data Analysis and Interpretation, Methodology, Software, Investigation, Validation, Evaluation, Formal Analysis, Visualization, Writing - Original Draft, Writing - Review \& Editing, Challenge Organization, Resources.\vspace{12pt}

\noindent{\bf D. Alapatt} : Conceptualization, Investigation, Validation, Evaluation, Formal Analysis, Visualization, Writing - Original Draft, Writing - Review \& Editing, Challenge Organization, Resources.\vspace{12pt}

\noindent{\bf T. Yu} : Data Curation, Investigation, Formal Analysis, Writing - Original Draft, Writing - Review \& Editing, Visualization.\vspace{12pt}

\noindent{\bf A. Vardazaryan} : Conceptualization, Investigation, Validation, Writing - Review \& Editing, Challenge Organization, Resources.\vspace{12pt}

\noindent{\bf F. Xia, Z. Zhao, T. Xia, F. Jia, Y. Yang, H. Wang, D. Yu, G. Zheng, X. Duan, N. Getty, R. Sanchez-Matilla, M. Robu, L. Zhang, H. Chen, J. Wang, B. Zhang, B. Gerats, S. Raviteja, R. Sathish, R. Tao, S. Kondo, W. Pang, H. Ren, J.R. Abbing, M.H. Sarhan, S. Bodenstedt, N. Bhasker, B. Oliveira, H. Torres, L. Ling, F. Gaida, T. Czempiel, Y. Jiang, Y. Long, J. Vila\c{c}a, P. Morais, J. Fonseca, R.M. Egging, I.N. Wijma, C. Qian, G. Bian, Z. Li, V. Balasubramanian, D. Sheet, I. Luengo, Y. Zhu, S. Ding, J. Aschenbrenner, N.E van der Kar, M. Xu, M. Islam, L. Seenivasan, A.C. Jenke, D. Stoyanov} : Methodology, Software, Writing - Review \& Editing.\vspace{12pt}

\noindent{\bf C. Gonzalez, B. Seeliger, P. Mascagni} : Data Curation, Writing - Review \& Editing.\vspace{12pt}

\noindent{\bf D. Mutter} : Data Curation, Writing - Review \& Editing, Supervision.\vspace{12pt}

\noindent{\bf N. Padoy} : Conceptualization, Writing - Review \& Editing, Supervision, Challenge Organization, Resources, Funding Acquisition, Project Administration.